%% file: main.tex
\pdfoutput=1

\documentclass[11pt]{article}

\usepackage[final]{acl}

\usepackage{times}
\usepackage{latexsym}

\usepackage[T1]{fontenc}

\usepackage[utf8]{inputenc}

\usepackage{microtype}

\usepackage{inconsolata}

\usepackage{graphicx}
\usepackage{adjustbox}
\usepackage{hhline}
\usepackage{xcolor,colortbl}
\usepackage{multirow}
\usepackage{booktabs}
\usepackage{url}
\usepackage[colorinlistoftodos]{todonotes}
\usepackage{amssymb}
\usepackage{tcolorbox}
\usepackage{float}
\usepackage{subcaption}
\usepackage{placeins}

\usepackage{amsmath}
\usepackage{xspace}
\newcommand{\task}{downstream task\xspace}
\newcommand{\tasks}{downstream tasks\xspace}
\newcommand{\basemodel}{aligned model\xspace}
\newcommand{\basemodels}{aligned models\xspace}
\title{Safeguard Fine-Tuned LLMs Through Pre- and Post-Tuning \\ Model Merging}

\author{
    \textbf{Hua Farn}$^\heartsuit$ \qquad
    \textbf{Hsuan Su}$^\heartsuit$ \qquad
    \textbf{Shachi H Kumar}$^\diamondsuit$ \qquad
    \\
    \textbf{Saurav Sahay}$^\diamondsuit$ \qquad
    \textbf{Shang-Tse Chen}$^\heartsuit$ \qquad
    \textbf{Hung-yi Lee}$^\heartsuit$ \qquad
    \\
    $^\heartsuit$National Taiwan University \qquad
    $^\diamondsuit$Intel Lab \qquad
    \\
    \small \texttt{alhena.farn@gmail.com}
}

\begin{document}
\maketitle

\input{sections/abstract}

\input{sections/introduction}

\input{sections/related_work}

\input{sections/method}

\input{sections/experiment_setup}

\input{sections/result}
\input{sections/conclusion}
\input{sections/limitation}

\bibliography{custom}

\appendix
\input{sections/appendix}

\end{document}

%% file: sections/abstract.tex
\begin{abstract}
Fine-tuning large language models (LLMs) for \tasks often leads to catastrophic forgetting, notably degrading the safety of originally aligned models. While some existing methods attempt to restore safety by incorporating additional safety data, the quality of such data typically falls short of that used in the original alignment process. Moreover, these high-quality safety data are generally inaccessible, making it difficult to fully recover the model's original safety. We ask: \textit{How can we preserve safety while improving \task performance \textbf{without} additional safety data?} We show that simply merging the weights of pre- and post-fine-tuned models effectively mitigates safety degradation while enhancing performance. Experiments across different \tasks and models validate the method’s practicality and effectiveness.
\end{abstract}

%% file: sections/introduction.tex
\section{Introduction}
The rapid advancement and increasing accessibility of Large Language Models (LLMs) necessitate a critical focus on aligning these technologies with human values, cultural norms, and trustworthiness \cite{huang2023surveysafetytrustworthinesslarge}. To address these challenges, researchers and developers have introduced safety techniques such as preference tuning \cite{ouyang2022traininglanguagemodelsfollow,dpo, grattafiori2024llama3herdmodels,openai2024gpt4technicalreport}, aimed at preventing LLMs from generating harmful or inappropriate content. 
Many applications now leverage safety-aligned models as foundation models—referred to as \textit{\basemodels} in this paper—to further customize for \tasks via supervised fine-tuning (SFT) \cite{sft}.
\begin{figure}[tp!]
    \begin{adjustbox}{width=\linewidth}
    \centering
    \includegraphics{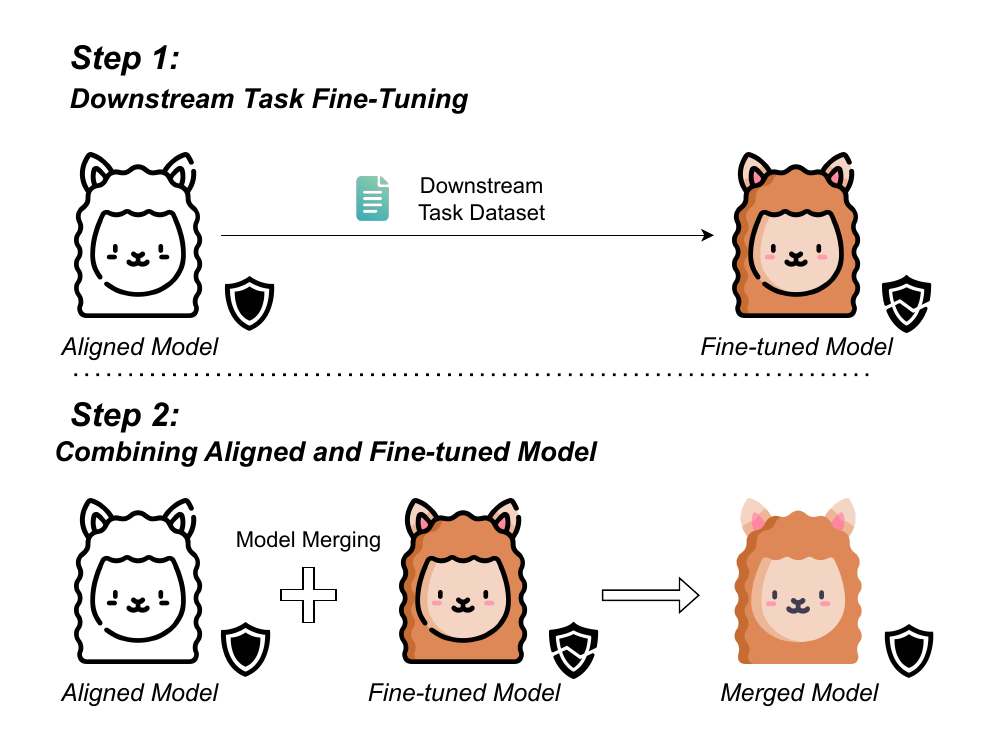}
    \end{adjustbox}
    \caption{Beyond standard SFT for \task adaptation, we can effectively mitigates safety degradation by combining the aligned and the fine-tuned model.}
    \vspace{-2mm}
    \label{fig:overview}
\end{figure}

However, recent studies \cite{yang2023shadowalignmenteasesubverting, qi2024finetuning, zhan-etal-2024-removing} highlight a critical challenge: fine-tuning \basemodels can degrade their safety, even when using benign datasets. To address this issue, mainstream approaches often incorporate additional safety data during fine-tuning \cite{qi2024finetuning, bianchi2024safetytuned}. However, since the original safety data used to align LLMs are rarely available, surrogate data are typically generated by other LLMs-raising concerns about quality, and the potential for alignment drift.


In this paper, we demonstrate a simple yet effective method for improving \task performance while mitigating safety degradation.
As illustrated in Figure~\ref{fig:overview}, our approach consists of two steps: (1) fine-tune the \basemodel on the \task, and (2) merge the \basemodel with the fine-tuned model. We evaluate this strategy across various models and \tasks. 
Experimental results show that this method consistently enhances \task performance while substantially preserving model safety, offering a simple and robust solution for fine-tuning safety-aligned LLMs.
Our key contributions are:
\begin{itemize}
    \item We show that a simple merging strategy can improve \task performance while lowering the Attack Success Rate (ASR).
    \item We conduct extensive evaluations across three LLMs, four \tasks, and two safety benchmarks, demonstrating the robustness of our method in preserving model safety.
\end{itemize}

%% file: sections/related_work.tex
\section{Related Work}

\subsection{Catastrophic Forgetting and Safety Degradation in LLMs}

LLMs are commonly aligned with human preferences to ensure safety and reduce the likelihood of generating harmful content \cite{ouyang2022traininglanguagemodelsfollow, dpo, grattafiori2024llama3herdmodels, openai2024gpt4technicalreport}. However, recent studies have shown that this safety alignment can be significantly compromised after fine-tuning on downstream tasks \cite{yang2023shadowalignmenteasesubverting, qi2024finetuning, zhan-etal-2024-removing}. This degradation is often attributed to catastrophic forgetting \cite{overcomingcatastrophic, li2024examiningforgettingcontinualpretraining,luo2025empiricalstudycatastrophicforgetting}, a well-known challenge in post-training scenarios where a model forgets previously acquired knowledge when adapting to new tasks.

To mitigate this issue, prior work has proposed several approaches. One line of work augments fine-tuning with additional safety data \cite{qi2024finetuning, bianchi2024safetytuned, safetyfinetuningatnocost}, aiming to reinforce desirable behaviors through curated examples. Another line of work leverages self-distillation, where the model generates synthetic training data, and fine-tuning on this data has been shown to reduce harmful tendencies \cite{selfdistill}.
In addition, some studies explore incorporating regularization strategies during training, often combined with additional safety data, to constrain harmful deviations \cite{huang2024lisa, huang2024vaccine}. Others adopt post-hoc re-alignment methods, such as \citet{antidote} and \citet{yi2024safetyrealignmentframeworksubspaceoriented}, which utilize additional safety data to identify safety-related masks and subsequently apply these masks in their re-alignment processes.

However, these methods either require synthesizing large amounts of safety data or incur significant computational overhead. In contrast, the approach proposed in this paper avoids both additional data requirements and extra training costs, offering a more efficient alternative for maintaining safety.





\subsection{Model Merging}

Model merging combines multiple models into a single unified model. A straightforward approach is to average the weights of different models \cite{linearmerging}, while variant techniques include SLERP \cite{slerp} and DARE \cite{dare}.

Another line of work explores \textit{task vectors} \cite{taskvector}, typically computed as the difference between a fine-tuned model and its base. These vectors enable composable transformations across tasks \cite{chatvector, syn2real} and have been extended to construct “safety vectors” from separate safe or harmful models to prevent safety degradation. 
\citet{homersimpson}, \citet{safetyarithmetic}, and \citet{seperatethewheat} adopt a similar approach: they first fine-tune an aligned model on harmful data to obtain a harmful variant, then compute a safety vector as the parameter difference between the aligned and harmful models, capturing the directional shift introduced by safety alignment. In contrast, \citet{safelora} avoids additional fine-tuning but assumes access to a pre-alignment checkpoint to derive the safety vector for subsequent alignment, which is not always publicly available.
In contrast, our method uses only aligned models and fine-tuned models, making it widely applicable, and demonstrates that safety 
can be restored without extra safety data.

The proposed approach is similar to WiSE-FT \cite{wortsman2022robust}, which also interpolates between the base model and its fine-tuned variant. However, WiSE-FT is applied to computer vision, not LLMs, and is not aimed at preserving safety.


%% file: sections/method.tex
\section{Methodology}

Our method comprises just two stages: (1) fine-tuning the \basemodel on a target \task, and (2) merging the original \basemodel with the fine-tuned model by interpolating their weights. Despite its simplicity, this merging strategy effectively mitigates the degradation in safety commonly observed following fine-tuning, while preserving performance on the target \task, without requiring additional data.

\paragraph{Step 1: Supervised Fine-Tuning of the Large Language Model}
We fine-tune the \basemodel with parameters $\theta_{\text{base}}$ on a given task $t$, resulting in a task-specific model $\theta_t$. For each task $t$, given an instruction $x^t$ and its corresponding response $y^t$, we minimize the negative log-likelihood:
\[
\mathcal{L}_{FT} = -\log f_{\theta}(y^t \mid x^t) \tag{1}
\]
where $f_\theta$ denotes the language model parameterized by $\theta$.

\paragraph{Step 2: Merging the Fine-Tuned Model with the Aligned Model}
After fine-tuning, we merge the parameters of the \basemodel ($\theta_{\text{base}}$) with those of the fine-tuned model ($\theta_{t}$) via linear interpolation:
\[
\theta_{\text{merged}} = (1 - \lambda) \theta_{\text{base}} + \lambda \theta_{t} \tag{2} \label{eq2}
\]
Here, $\theta_{\text{merged}}$ denotes the parameters of the merged model, and $\lambda \in [0, 1]$ controls the relative contribution of the fine-tuned model.
Eq.~\ref{eq2} is the formulation for the native linear merging method; other advanced merging methods can also be applied.

%% file: sections/experiment_setup.tex
\section{Experimental Setups}


\paragraph{Downstream Tasks}
We conduct experiments on four \tasks: reasoning, medical assistance, code generation, and tool usage proficiency.
Reasoning is enhanced using Chain-of-Thought data from the Flan V2 dataset \cite{longpre2023flancollectiondesigningdata} and evaluated on the Big Bench Hard (BBH) dataset \cite{BBH}. 
Medical assistance uses patient-doctor dialogues from the ChatDoctor dataset \cite{li2023chatdoctormedicalchatmodel}. 
Code generation is trained on the MagiCoder dataset \cite{wei2024magicoder} and evaluated using the HumanEval benchmark \cite{humaneval}. 
Tool usage proficiency leverages the OpenFunctions dataset \cite{openfunction} to improve API call generation. For medical assistance and tool usage proficiency, response similarity to reference answers is measured using BERTScore\footnote{Embeddings extracted from the 40th layer of \texttt{microsoft/deberta-xlarge-mnli}.} \cite{Zhang*2020BERTScore:}. 
See Appendix~\ref{Appendix:Task Domain-Specific Tasks Detail} for additional details on the \tasks.

\paragraph{Safety Evaluation}
We assess safety using harmful instructions from the AdvBench \cite{advbench} and HEx-PHI \cite{qi2024finetuning} datasets. 
Following prior works that use safety classifiers to automatically detect harmful content \cite{xie2025sorrybench, obrien2024steeringlanguagemodelrefusal}, 
we adopt WildGuard \cite{han2024wildguard}, a classifier shown to perform comparably to GPT-4 \cite{openai2024gpt4technicalreport}. 
We report the Attack Success Rate (ASR) as the primary evaluation metric. 
Details of the evaluation setup are provided in Appendix~\ref{Appendix: safety benchmark and classifier}.

\paragraph{Large Language Models}  
Our experiments involve several LLMs, including LLaMA-3-8B-Instruct \cite{grattafiori2024llama3herdmodels}, Gemma-2-2B-It \cite{gemmateam2024gemmaopenmodelsbased}, and Qwen2.5-7B-Instruct \cite{qwen2.5}, along with additional model sizes when noted. 
We use the \textit{instruct-tuned} variants of all models, which are aligned with human preferences. 
Each model is fine-tuned on each \task using LoRA \cite{hu2022lora} with three different random seeds. 
The reported \task performance and ASR are averaged across these three runs. 
Additional details of experiment are provided in Appendix~\ref{Appendix:Experimental Details}.

\paragraph{Baselines}
Unlike most existing methods aimed at mitigating safety degradation in LLMs after fine-tuning, our proposed approach requires neither additional data nor further training. Given the absence of comparable safety alignment techniques, we evaluate our method’s efficacy in preserving the safety attributes of the originally aligned model post fine-tuning by benchmarking it against two prevalent regularization techniques: Dropout \cite{dropout} and Weight Decay \cite{weightdecay}. Similar to our approach, these regularization methods do not necessitate extra data or further training. The hyperparameters for these techniques are selected based on validation set performance on \tasks. 

\paragraph{Merging Methods}
In Section~\ref{results and discussion}, we used Linear Merging, which combines models via direct interpolation as defined in Eq.~\ref{eq2}, as the merging method. Two advanced merging methods—SLERP and DARE—are also applied. Their results are provided in Appendix~\ref{Appendix: More Results}. For all methods, we merge each fine-tuned model with the \basemodel using an interpolation factor $\lambda$ selected based on validation set performance. 



%% file: sections/result.tex
\section{Results}
\label{results and discussion}

\subsection{Can model merging mitigate safety degradation after fine-tuning?}
\label{Results: Q1}
\input{figures_tex/pareto_front_advbench}

Figure~\ref{fig:pareto_front_advbench} presents a Pareto analysis of task performance and ASR on AdvBench across different models and tasks. 
We observe that SFT consistently leads to safety degradation, with higher ASR across all settings compared to the original \basemodel. While Dropout and Weight Decay offer slight improvements in ASR, they are generally insufficient to restore the safety of the \basemodel.

In contrast, the proposed approach consistently achieves a better balance between performance and safety. It often reduces ASR to levels near that of the \basemodel while maintaining—or even improving—task performance. The smooth Pareto fronts formed by merging indicate controllable trade-offs, making it an effective solution for mitigating safety loss after fine-tuning. Figure \ref{fig:pareto_front_hexphi} shows the HEx-PHI results, and results of different merging methods can be found in Table \ref{tab:main} in Appendix \ref{Appendix: More Results}.

\subsection{How does model merging perform across different model sizes?}
\input{figures_tex/diff_size}
\input{figures_tex/ifeval_llama3}

\citet{luo2025empiricalstudycatastrophicforgetting} noted that larger models may suffer more from catastrophic forgetting. We extend this analysis to safety degradation and evaluate how model merging performs across different model sizes.
Figure~\ref{fig:diff_size} shows the average changes in performance and ASR across all \tasks for the Qwen2.5 and Gemma-2 model families, comparing SFT and the proposed approach against their respective \basemodels. Both methods improve task performance, with larger models generally achieving greater gains. However, safety degradation shows no consistent trend: smaller Qwen models degrade more, while larger Gemma models are more affected. This suggests that safety degradation is not solely determined by model size. Nonetheless, the proposed approach consistently mitigates safety degradation across different scales.

\subsection{Can model merging help preserve other capabilities of the \basemodel?}
While our method mitigates safety degradation, we also investigate whether it preserves other capabilities of the \basemodel that are lost due to catastrophic forgetting. Since we fine-tune \textit{instruct-tuned} variants, we evaluate whether instruction-following ability is retained.
Figure~\ref{fig:ifeval_llama3} shows the performance of LLaMA-3-8B-Instruct on the instruction-following benchmark IFEval \cite{ifeval}. Both prompt and instruction accuracy decline after fine-tuning, with the largest drops observed in the reasoning and medical tasks. The proposed approach substantially restores performance to the level of the \basemodel, indicating that merging can also preserve instruction adherence.

%% file: figures_tex/pareto_front_advbench.tex
\begin{figure*}[ht!]
    \centering
    \includegraphics[width=\textwidth]{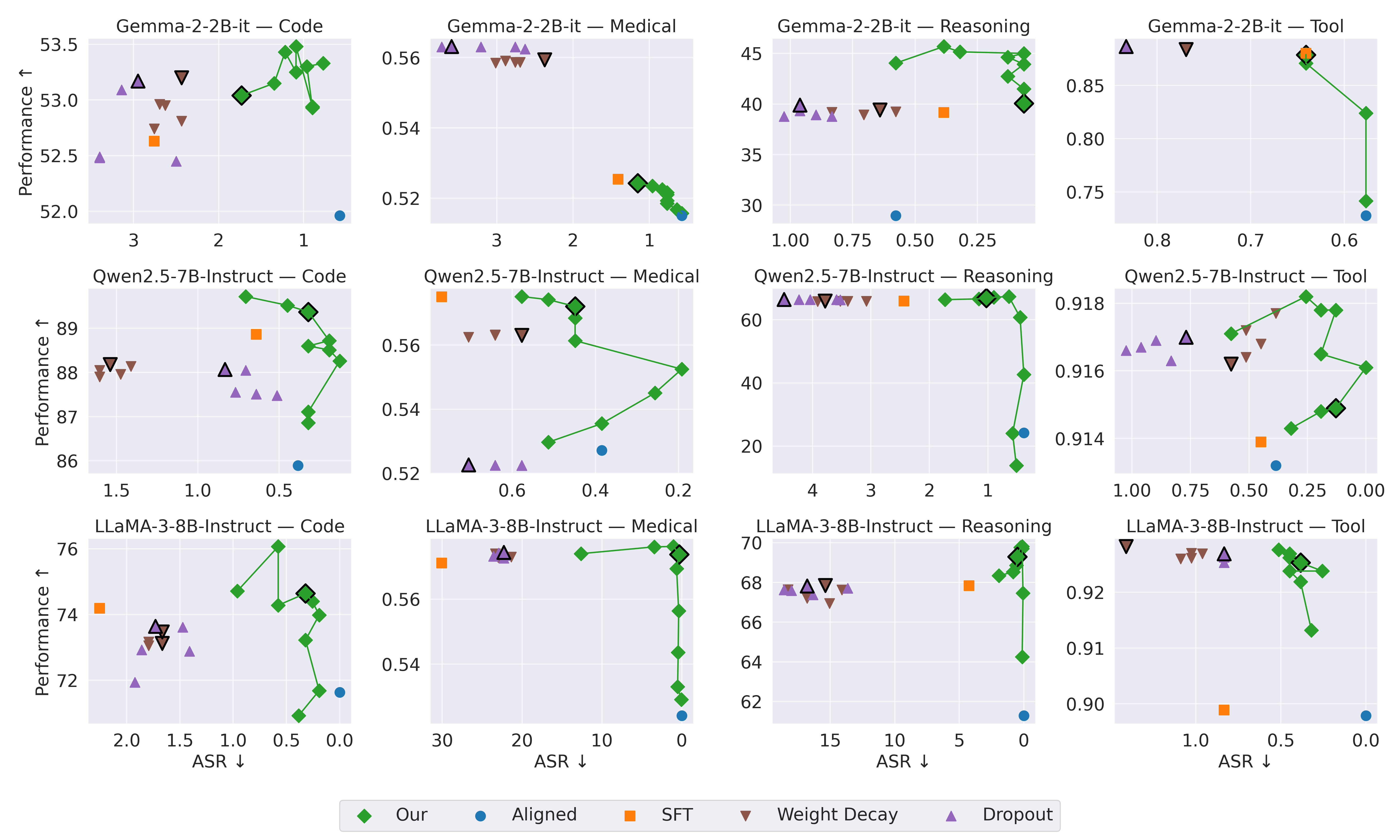}
    \caption{\textbf{Pareto analysis of \task performance and ASR on AdvBench across different models and tasks.} 
    Each dot represents a model configuration, with different hyperparameter settings (weight decay coefficient, dropout rate, or merging interpolation coefficient) for the same method shown in the same color. For clarity, we connect the dots of our method in ascending order of their coefficients. 
    Dots with dark edges indicate the best-performing models on the validation set for each method.}

    \label{fig:pareto_front_advbench}
    \vspace{-2mm}
\end{figure*}

%% file: figures_tex/diff_size.tex
\begin{figure}[ht!]
    \captionsetup{aboveskip=2pt, belowskip=2pt} 
    \centering
    \includegraphics[width=\linewidth]{figures/QwenAvgen_all.png}
    \vspace{5mm} 
    \includegraphics[width=\linewidth]{figures/GemmaAvgen_all.png}
    \caption{
\textbf{Performance and ASR change across model sizes.}
This figure shows results for Qwen2.5 at 1.5B, 3B, and 7B (top), and Gemma-2 at 2B and 9B (bottom).
}

    \label{fig:diff_size}
    \vspace{-2mm}
\end{figure}



%% file: figures_tex/ifeval_llama3.tex
\begin{figure}[ht!]
    \centering
    \includegraphics[width=\linewidth]{figures/ifeval_llama3.png}
    \caption{
\textbf{Accuracy of LLaMA-3 on IFEval.} This figure shows results of LLaMA-3-8B-Instruct fine-tuned on downstream tasks on IFEval. Fine-tuning reduces instruction-following ability, especially for Reasoning and Medical tasks. Merging with the Aligned model helps recover this ability close to the original level.
}

    \label{fig:ifeval_llama3}
    \vspace{-2mm}
\end{figure}

%% file: sections/conclusion.tex
\section{Conclusion}
We present a simple yet effective method to address the safety degradation that often occurs when adapting LLMs to \tasks, without requiring additional safety data or auxiliary models. The method also preserves capabilities such as instruction-following, making it a practical and scalable solution for adapting LLMs to new tasks.

%% file: sections/limitation.tex
\section{Limitations}

\paragraph{Task and Model Selection}
In our experiments, we evaluate only on benign data from four task domains: reasoning, medical assistance, code generation, and tool-using proficiency. Other important areas such as law, finance, or multilingual tasks remain unexplored. While Section~\ref{results and discussion} shows the effectiveness of our method on the selected \tasks, its generalizability to other domains, languages, or datasets that may contain harmful content remains an open question. Additionally, we evaluate models with sizes ranging from 1.5B to 9B across three model families. The effectiveness of our approach on larger models or different model architectures warrants further investigation.

\paragraph{Safety Classifier for Safety Evaluation}
Due to the high computational and financial cost of human-aligned safety evaluation methods such as LLM-as-Judge \cite{chiang-lee-2023-large, geval}, which require using large proprietary models like GPT-4 \cite{openai2024gpt4technicalreport}, we instead adopt WildGuard \cite{han2024wildguard}, a lightweight open-source safety classifier. WildGuard is shown to perform competitively with GPT-4 on multiple safety detection tasks and offers a reproducible, low-cost alternative suitable for large-scale evaluations.

However, this classifier-based approach has several limitations. First, WildGuard may struggle with complex or subtle harmful instructions, potentially leading to both false positives and false negatives. Second, it provides only binary or coarse-grained outputs (e.g., “harmful” or “safe”), without offering finer distinctions such as the category of harm, the severity of the risk, or whether the model’s refusal was appropriate or evasive.

Consequently, while WildGuard enables efficient and scalable evaluation, it constrains the depth of our safety analysis. Future work could incorporate more fine-grained multi-label safety classifiers, adversarial evaluation pipelines, or hybrid setups involving human or LLM-as-Judge verification to better capture the nuanced impact of model merging on safety behavior.

\paragraph{Jailbreak Attacks}
Our work focuses on safety degradation that arises from fine-tuning aligned LLMs on benign tasks, which we consider a case of catastrophic forgetting. As such, we evaluate whether models produce harmful outputs when directly prompted with harmful instructions, rather than testing resistance to specific jailbreak strategies. We do not include jailbreak-style attacks \cite{jailbreaksurvey} in our evaluation due to two reasons: (1) Our primary goal is to study alignment loss under standard fine-tuning, not adversarial robustness; and (2) jailbreak evaluations typically require separate prompting strategies and adversarial instruction crafting pipelines, which are beyond the scope of this study. Future work can extend our framework to examine the impact of merging on robustness against jailbreak attacks.

\section{Ethics Statement}

While our method effectively addresses safety degradation in aligned LLMs without requiring additional safety data, our approach relies on merging pre- and post-fine-tuned models to preserve safety, which may inadvertently inherit any latent biases or unsafe behaviors that are still presented in the base model. Further investigation is needed to explore the impact of these inherited biases in the base model.

%% file: sections/appendix.tex
\section{Domain-Specific Tasks Detail}
\label{Appendix:Task Domain-Specific Tasks Detail}
\paragraph{\textbf{Reasoning}}
We randomly select 10,000 zero-shot chain-of-thought instructions from the Flan V2 dataset then split them into training set and validation set with ratio $9:1$. Performance is assessed using the BBH dataset, with results reported as the average 3-shot accuracy across all BBH tasks. We use lm-evaluation-harness \cite{eval-harness} as our code base.
\paragraph{Medical Assistance}
We randomly select 10,000 real patient-doctor conversations from the ChatDoctor dataset \cite{li2023chatdoctormedicalchatmodel} then split them into training set and validation set with ratio $9:1$. Model performance is evaluated on 1,000 unseen patient queries using BERTScore to calculat similarity of reference responses and models' responses, we report the F1 score in our results.
\paragraph{Code Generation}
We select 10,000 samples from the MagiCoder dataset \cite{wei2024magicoder} to improve code generation capabilities. Specifically, we uniformly sampled from each coding languages. When evaluating on HumanEval, we set $n = 50$, representing the number of responses generated per question, and report Pass@10 as our evaluation metric. During evaluation, we prepend the instruction: \textit{"Complete the following code and return only the completed code, without any explanations or additional text."} to enforce that the model generates only executable code.
\paragraph{Tool Using Proficiency}
Due to the smaller size of the OpenFunctions dataset \cite{openfunction}, we split its full training set into training and validation subsets using a 9:1 ratio to enhance the model's API call generation capabilities. The model is evaluated on the full OpenFunctions test set, with performance measured using BERTScore to compute the similarity between the reference responses and the model outputs. We report the F1 score as our evaluation metric. During evaluation, we prepend the instruction: \textit{"Complete the following code and return only the completed code, without any explanations or additional text."} to ensure that the model generates only executable function calls.
\paragraph{Instruction Following}
To assess whether instruction-following ability is preserved after fine-tuning, we evaluate models on IFEval \cite{ifeval}, a benchmark specifically designed to test instruction adherence. We use the official IFEval evaluation set and report both prompt and instruction accuracy. Since our models are fine-tuned from \textit{instruct-tuned} variants, this evaluation helps determine whether merging can recover instruction-following capabilities degraded during task-specific fine-tuning. We use \texttt{lm-evaluation-harness} \cite{eval-harness} as the code base for this evaluation.

\section{Safety Dataset and Classifier}
\label{Appendix: safety benchmark and classifier}
\subsection{Safety Dataset}
\label{Appendix: Safety Benchmark}

\paragraph{AdvBench}\cite{advbench} is a set of 520 harmful behaviors formulated as instructions. The goal of these instructions is to identify a single attack string that causes the model to generate responses compliant with the instruction across as many harmful behaviors as possible.

\paragraph{HEx-PHI} \cite{qi2024finetuning} is a dataset containing 330 harmful instructions designed to evaluate the harmfulness of LLMs. It includes 30 examples for each of 11 prohibited categories, which are derived from a combination of OpenAI’s usage policies and Meta’s Llama 2 acceptable use policy. These categories are:

\begin{enumerate}
    \item Illegal Activity
    \item Child Abuse Content
    \item Hate / Harassment / Violence
    \item Malware
    \item Physical Harm
    \item Economic Harm
    \item Fraud / Deception
    \item Adult Content
    \item Political Campaigning
    \item Privacy Violation Activity
    \item Tailored Financial Advice
\end{enumerate}

\subsection{Safety Classifier}
\label{Appendix: Safety Classifier}
WildGuard \cite{han2024wildguard} is an open, one-stop moderation model that achieves three goals: 
1) Detection of harm in user prompts, 
2) Detection of harm in LLM responses, and 
3) Refusal evaluation of LLM responses. Comprehensive evaluations on WildGuardTest \cite{han2024wildguard} and ten existing public benchmarks that are shown in \citet{han2024wildguard} demonstrate that WildGuard outperforms the strongest existing open-source baselines in F1 scores and matches GPT-4 across various tasks.

When evaluating the responses generated by LLMs to harmful instructions, we apply the recommended settings, such as the instruction format and decoding strategy of WildGuard\footnote{\url{https://huggingface.co/allenai/wildguard}}. After evaluation, WildGuard outputs the response in the following format:
\begin{tcolorbox}[width=\columnwidth,colback=white]
\small
\begin{verbatim}
Harmful request: yes/no
Response refusal: yes/no
Harmful response: yes/no
\end{verbatim}
\end{tcolorbox}
We then parse the result of the harmful response. If the result cannot be parsed, we count it as a miss. Our final ASR is calculated as follows:
\[
ASR = \frac{N_{yes}}{N_{total} - N_{miss}}\times 100\%
\]
where $N_{yes}$ is the number of harmful responses classified as "yes", $N_{total}$ is the total number of responses, and $N_{miss}$ is the number of responses that failed to be parsed. In our experiments, $N_{miss}$ is negligible for all tested models across both safety datasets.

\section{Experimental Detail}
\label{Appendix:Experimental Details}
\subsection{Prompt Template}

For \basemodels, we directly apply their own prompt templates during the training and inference phases. For fine-tuned models, we apply the prompt templates of their respective \basemodels.

For the Llama-3 family, we use the following prompt template with a system prompt: \texttt{You are a helpful assistant.} for the tasks of reasoning, code generation, and tool usage proficiency:

\begin{tcolorbox}[width=\columnwidth,colback=white,boxrule=0.5mm]
\tiny
\begin{verbatim}
<|begin_of_text|><|start_header_id|>system<|end_header_id|>

You are a helpful assistant.<|eot_id|>

<|start_header_id|>user<|end_header_id|>

{Instruction}<|eot_id|>

<|start_header_id|>assistant<|end_header_id|>

{Response}
\end{verbatim}
\end{tcolorbox}

For the medical assistance task, we use another prompt provided in the ChatDoctor dataset \cite{li2023chatdoctormedicalchatmodel} as the system prompt. Hence, the prompt is as follows:

\begin{tcolorbox}[width=\columnwidth,colback=white,boxrule=0.5mm]
\tiny
\begin{verbatim}
<|begin_of_text|><|start_header_id|>system<|end_header_id|>

If you are a doctor, please answer the medical 
questions based on the patient's description.<|eot_id|>

<|start_header_id|>user<|end_header_id|>

{Instruction}<|eot_id|>

<|start_header_id|>assistant<|end_header_id|>

{Response}
\end{verbatim}
\end{tcolorbox}

The prompt for Gemma-2 for the tasks of reasoning, code generation, and tool usage proficiency is shown below:

\begin{tcolorbox}[width=\columnwidth,colback=white,boxrule=0.5mm]
\tiny
\begin{verbatim}
<bos><start_of_turn>user
You are a helpful assistant.{Instruction}<end_of_turn>
<start_of_turn>model
{Response}
\end{verbatim}
\end{tcolorbox}

The prompt for the medical assistance task is as follows:

\begin{tcolorbox}[width=\columnwidth,colback=white,boxrule=0.5mm]
\tiny
\begin{verbatim}
<bos><start_of_turn>user
If you are a doctor, please answer the medical 
questions based on the patient's description.
{Instruction}<end_of_turn>
<start_of_turn>model
{Response}
\end{verbatim}
\end{tcolorbox}

The prompt for Qwen2.5 for the tasks of reasoning, code generation, and tool usage proficiency is shown below:
\begin{tcolorbox}[width=\columnwidth,colback=white,boxrule=0.5mm]
\tiny
\begin{verbatim}
<|im_start|>system
You are a helpful assistant.
<|im_end|>
<|im_start|>user
{Instruction}
<|im_end|>
<|im_start|>assistant
{Response}
\end{verbatim}
\end{tcolorbox}

The prompt for the medical assistance task is as follows:

\begin{tcolorbox}[width=\columnwidth,colback=white,boxrule=0.5mm]
\tiny
\begin{verbatim}
<|im_start|>system
If you are a doctor, please answer the medical 
questions based on the patient's description.
<|im_end|>
<|im_start|>user
{Instruction}
<|im_end|>
<|im_start|>assistant
{Response}
\end{verbatim}
\end{tcolorbox}

\subsection{Fine-tuning}
\label{Appendix: Fine-tuning}

For all tasks, we fine-tune three model instances using different random seeds: 42, 1024, and 48763. We employ LoRA with $r=8$ and $\alpha=16$ for all linear layers, utilizing the AdamW optimizer with a learning rate of $1 \times 10^{-4}$ and a cosine learning rate scheduler. We use a batch size of 8 and train for 3 epochs. All models are trained on either an RTX A6000 GPU or an RTX 6000 Ada Generation GPU using LLaMA-Factory \cite{zheng2024llamafactory} as the codebase.

Although we initially fine-tuned each task for 3 epochs, we observed stronger model performance at an earlier stage. Consequently, unless explicitly stated otherwise, we report model training after 500 steps for reasoning, medical assistance, and code generation, and after 200 steps for tool usage proficiency due to the smaller size of the OpenFunctions training set.
\subsection{Baseline Methods}
\label{Appendix: Baseline}
We evaluate dropout rates in the range of $0.1$ to $0.5$, and weight decay coefficients also from $0.1$ to $0.5$. The optimal hyperparameters for each technique are selected based on performance on the \tasks validation set.
\subsection{Inference}
\label{Appendix: Inference}

We use greedy decoding to ensure result consistency, except for the HumanEval benchmark. For HumanEval, we apply sampling-based decoding with a temperature of 0.6, top\_p of 0.9, top\_k of 50, and a repetition penalty of 1.2. To accelerate the inference process, we utilize the vLLM engine \cite{vllm} for model inference.

\section{Model Merging}
\label{Appendix: Merging}
\subsection{Merging Methods}
\label{Appendix: Merging Methods}

\paragraph{Linear Merging}
Linear Merging involves directly combining the weights of the \basemodel and the fine-tuned model by interpolating their parameters. Specifically, the weights of the merged model are calculated as a weighted average of the base and fine-tuned models' weights, following Equation \ref{eq2}. This method is straightforward and computationally efficient, making it a popular choice for basic model integration.

\paragraph{SLERP}
Spherical Linear Interpolation (SLERP) \cite{slerp} is an advanced merging technique that interpolates between model weights on a hypersphere, ensuring a smoother and more natural transition between the models. Unlike Linear Merging, SLERP accounts for the angular relationship between weight vectors, which aim to better preserve the \basemodel's features while effectively integrating the fine-tuned model's task-specific enhancements.

\paragraph{DARE}
Drop and Rescale (DARE) \cite{dare} is a method used to prepare models for merging techniques such as Linear Merging. It operates by randomly dropping parameters according to a specified drop rate and rescaling the remaining parameters. This process helps reduce the number of redundant and potentially interfering parameters among multiple models.

\subsection{Model Merging Implementation}
\label{Appendix:Model Merging Implementation}
We adopt MergeKit \cite{mergekit} as our implementation framework and only vary the interpolation factor $\lambda$. For Linear Merging, we test $\lambda$ values in the range ${0.1, 0.2, \dots, 0.9}$ with a step size of 0.1. For SLERP and DARE, we use the same range of $\lambda$ values and follow their respective default configurations in MergeKit—specifically, the default dot product threshold for SLERP and the default drop rate for DARE. 
\section{More Results}
\label{Appendix: More Results}

\subsection{Comparison of Different Methods}
\input{figures_tex/pareto_front_hexphi}

\input{tables/main_result}

In Section~\ref{Results: Q1}, we demonstrate that Linear Merging consistently achieves a better trade-off between performance and safety when evaluated on various \tasks and AdvBench. Figure~\ref{fig:pareto_front_hexphi} further confirms this trend on the HEx-PHI benchmark, where Linear Merging yields favorable Pareto fronts across different models and tasks.

To better reflect practical usage scenarios, we additionally report results based on the best-performing model (on the validation set of each task) within each method category—including Weight Decay, Dropout, Linear Merging, DARE, and SLERP. These results are summarized in Table~\ref{tab:main}, providing a fair comparison of each method’s effectiveness under optimal conditions. We use validation set performance for model selection, as it is commonly available during deployment and serves as a realistic basis for method comparison.

In Table~\ref{tab:main}, even when each method is allowed to select its best-performing checkpoint, merging-based approaches still exhibit strong capability in recovering the safety of the fine-tuned model, often outperforming regularization-based methods such as Dropout and Weight Decay. This suggests that model merging is not only effective but also practical for mitigating safety degradation in real-world settings, even without access to additional safety data.

\subsection{Which safety category suffers the most from safety degradation?}

In this section, we investigate which categories in HEx-PHI are most affected by safety degradation. All categories are listed in Appendix~\ref{Appendix: Safety Benchmark}.

As observed in Section~\ref{Results: Q1}, LLaMA-3-8B-Instruct and Qwen2.5-7B-Instruct exhibit the most severe degradation on the Reasoning and Medical Assistance tasks. Therefore, we analyze their responses on the HEx-PHI benchmark to further understand which safety categories are most impacted.

The category distributions are shown in Figure~\ref{fig:hexphi-category-breakdown}. For LLaMA-3-8B-Instruct, the \basemodel only generates harmful responses in categories 4 (Malware), 9 (Political Campaigning), and 10 (Privacy Violation Activity). After fine-tuning, however, harmful responses increase across all categories, with categories 4, 7 (Fraud/Deception), and 9 exhibiting the most significant growth in both tasks. This demonstrates that safety degradation extends to fine-grained category levels, making it difficult to address safety concerns solely by modifying the model prior to fine-tuning.

Qwen2.5-7B-Instruct shows a slightly different trend. Its \basemodel generates harmful responses across more categories compared to LLaMA-3-8B-Instruct, and fine-tuning further aggravates these issues. However, both models share a common pattern: a large number of harmful responses appear in categories 7 and 9. This suggests that certain categories may be particularly vulnerable to safety degradation during fine-tuning, regardless of model architecture or \task.

After applying different merging methods, most harmful categories show a reduction in the number of harmful responses. However, the degree of improvement varies across merging strategies and tasks. For instance, Linear Merging performs best on LLaMA-3-8B-Instruct but not on Qwen2.5-7B-Instruct, and some categories do not benefit from merging at all. This indicates that no single method universally outperforms others in preserving safety across all harmful categories. 
\input{figures_tex/harm_cat}




%% file: figures_tex/pareto_front_hexphi.tex
\begin{figure*}[ht!]
    \centering
    \includegraphics[width=\textwidth]{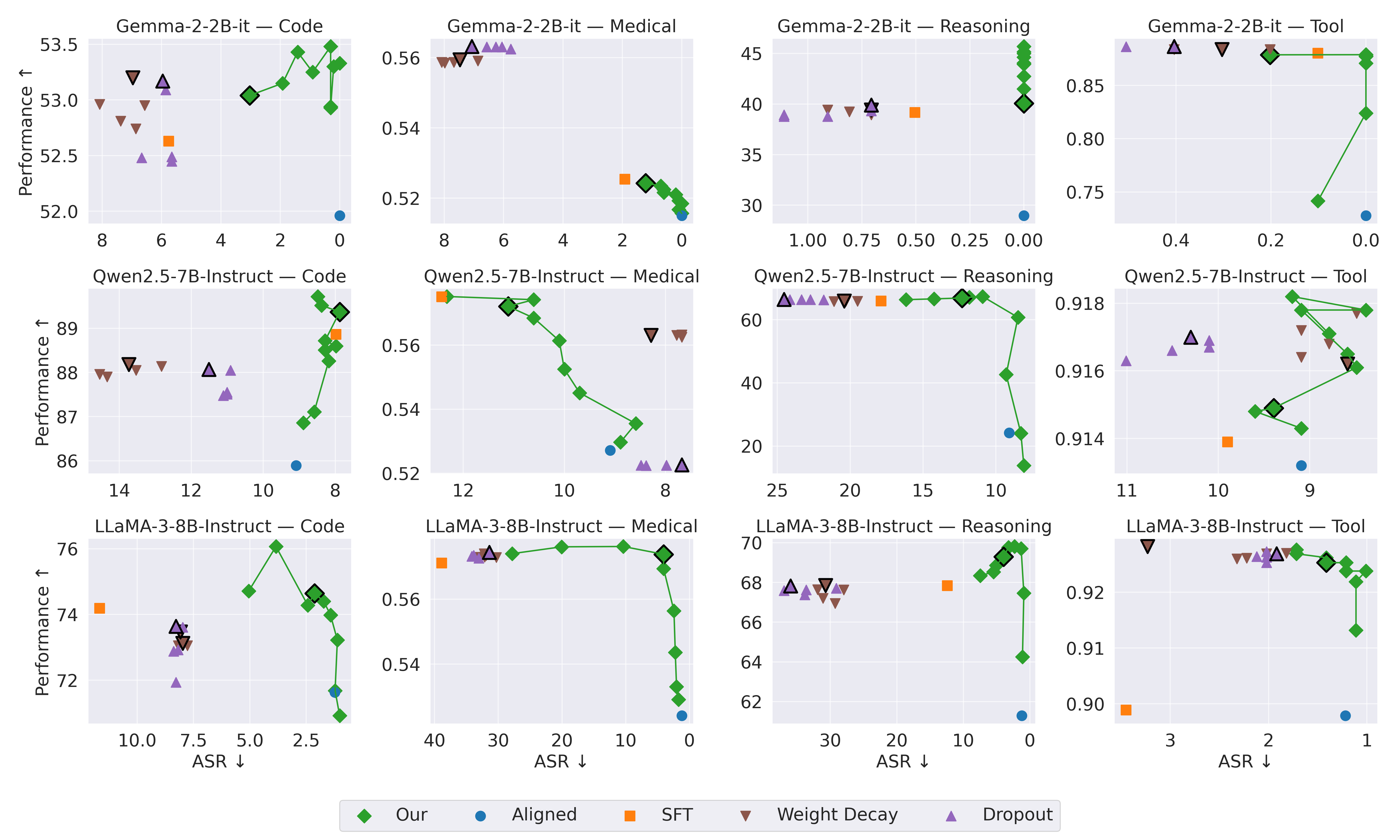}
    \caption{\textbf{Pareto analysis of \task performance and safety across different models and tasks.} We present the trade-off between performance and attack success rate (ASR) on HEx-PHI when applying weight decay, dropout, and Linear Merging.}

    \label{fig:pareto_front_hexphi}
    \vspace{-2mm}
\end{figure*}

%% file: tables/main_result.tex
\begin{table*}[ht]
\centering
\caption{\textbf{Performance and ASR on the \task.} We compare different merging methods with SFT and baselines. Merging often improves \task performance while retaining safety. Bold indicates the best score per metric (excluding \textbf{Aligned}). For the attack success rate on AdvBench and HEx-PHI, we report percentage values. For the downstream tasks of Reasoning and Code Generation, we report accuracy, while for the remaining two tasks, we report F1 scores.}

\begin{adjustbox}{width=\textwidth}
\renewcommand{\arraystretch}{1.15}
\begin{tabular}{cc|ccc|ccc|ccc}
\hline
\multirow{2}{*}{\textbf{Task}} & \multirow{2}{*}{\textbf{Method}} & \multicolumn{3}{c}{LLaMA-3-8B-Instruct} & \multicolumn{3}{c}{Gemma-2-2B-It} & \multicolumn{3}{c}{Qwen2.5-7B-Instruct} \\
\hhline{~~|---|---|---}
& & Perf. $\uparrow$ & AdvBench $\downarrow$ & HEx-PHI $\downarrow$ & Perf. $\uparrow$ & AdvBench $\downarrow$ & HEx-PHI $\downarrow$ & Perf. $\uparrow$ & AdvBench $\downarrow$ & HEx-PHI $\downarrow$ \\
\midrule

\multirow{7}{*}{Reasoning}
& Aligned & 61.30\% & 0.00\% & 1.22\% & 28.98\% & 0.58\% & 0.00\% & 24.16\% & 0.38\% & 9.09\% \\
\cmidrule{2-11}
& SFT & 67.84\% & 4.25\% & 12.41\% & 39.16\% & 0.38\% & 0.51\% & 65.94\% & 2.44\% & 17.88\% \\
& \cellcolor{green!20}Weight Decay & \cellcolor{green!20}67.85\% & \cellcolor{green!20}15.38\% & \cellcolor{green!20}30.71\% & \cellcolor{green!20}39.41\% & \cellcolor{green!20}0.19\% & \cellcolor{green!20}0.71\% & \cellcolor{green!20}65.92\% & \cellcolor{green!20}3.78\% & \cellcolor{green!20}20.40\% \\
& \cellcolor{green!20}Dropout & \cellcolor{green!20}67.83\% & \cellcolor{green!20}16.79\% & \cellcolor{green!20}35.96\% & \cellcolor{green!20}39.89\% & \cellcolor{green!20}0.96\% & \cellcolor{green!20}0.71\% & \cellcolor{green!20}66.45\% & \cellcolor{green!20}4.49\% & \cellcolor{green!20}24.55\% \\
& \cellcolor{cyan!20}Linear & \cellcolor{cyan!20}\textbf{69.23}\% & \cellcolor{cyan!20}\textbf{0.64}\% & \cellcolor{cyan!20}6.38\% & \cellcolor{cyan!20}\textbf{40.07}\% & \cellcolor{cyan!20}\textbf{0.06}\% & \cellcolor{cyan!20}\textbf{0.00}\% & \cellcolor{cyan!20}\textbf{66.96}\% & \cellcolor{cyan!20}1.03\% & \cellcolor{cyan!20}12.32\% \\
& \cellcolor{cyan!20}DARE & \cellcolor{cyan!20}68.64\% & \cellcolor{cyan!20}1.28\% & \cellcolor{cyan!20}\textbf{5.66}\% & \cellcolor{cyan!20}40.01\% & \cellcolor{cyan!20}0.10\% & \cellcolor{cyan!20}\textbf{0.00}\% & \cellcolor{cyan!20}66.89\% & \cellcolor{cyan!20}1.09\% & \cellcolor{cyan!20}\textbf{12.22}\% \\
& \cellcolor{cyan!20}SLERP & \cellcolor{cyan!20}68.68\% & \cellcolor{cyan!20}1.22\% & \cellcolor{cyan!20}5.86\% & \cellcolor{cyan!20}40.05\% & \cellcolor{cyan!20}0.26\% & \cellcolor{cyan!20}\textbf{0.00}\% & \cellcolor{cyan!20}66.73\% & \cellcolor{cyan!20}\textbf{0.96}\% & \cellcolor{cyan!20}13.03\% \\
\midrule

\multirow{7}{*}{Medical Assistance}
& Aligned & 0.5242 & 0.00\% & 1.22\% & 0.5151 & 0.58\% & 0.00\% & 0.5271 & 0.38\% & 9.09\% \\
\cmidrule{2-11}
& SFT & 0.5711 & 30.06\% & 38.85\% & 0.5254 & 1.41\% & 1.92\% & \textbf{0.5751} & 0.77\% & 12.42\% \\
& \cellcolor{green!20}Weight Decay & \cellcolor{green!20}0.5740 & \cellcolor{green!20}23.33\% & \cellcolor{green!20}32.22\% & \cellcolor{green!20}0.5594 & \cellcolor{green!20}2.37\% & \cellcolor{green!20}7.47\% & \cellcolor{green!20}0.5631 & \cellcolor{green!20}0.58\% & \cellcolor{green!20}8.28\% \\
& \cellcolor{green!20}Dropout & \cellcolor{green!20}0.5744 & \cellcolor{green!20}22.31\% & \cellcolor{green!20}31.41\% & \cellcolor{green!20}\textbf{0.5632} & \cellcolor{green!20}3.59\% & \cellcolor{green!20}7.07\% & \cellcolor{green!20}0.5226 & \cellcolor{green!20}0.71\% & \cellcolor{green!20}\textbf{7.68}\% \\
& \cellcolor{cyan!20}Linear & \cellcolor{cyan!20}0.5738 & \cellcolor{cyan!20}\textbf{0.32}\% & \cellcolor{cyan!20}\textbf{4.06}\% & \cellcolor{cyan!20}0.5243 & \cellcolor{cyan!20}\textbf{1.15}\% & \cellcolor{cyan!20}\textbf{1.21}\% & \cellcolor{cyan!20}0.5721 & \cellcolor{cyan!20}0.45\% & \cellcolor{cyan!20}11.11\% \\
& \cellcolor{cyan!20}DARE & \cellcolor{cyan!20}0.5758 & \cellcolor{cyan!20}5.61\% & \cellcolor{cyan!20}23.41\% & \cellcolor{cyan!20}0.5248 & \cellcolor{cyan!20}\textbf{1.15}\% & \cellcolor{cyan!20}\textbf{1.21}\% & \cellcolor{cyan!20}0.5724 & \cellcolor{cyan!20}\textbf{0.26}\% & \cellcolor{cyan!20}11.52\% \\
& \cellcolor{cyan!20}SLERP & \cellcolor{cyan!20}\textbf{0.5789} & \cellcolor{cyan!20}5.76\% & \cellcolor{cyan!20}24.26\% & \cellcolor{cyan!20}0.5243 & \cellcolor{cyan!20}\textbf{1.15}\% & \cellcolor{cyan!20}1.52\% & \cellcolor{cyan!20}0.5729 & \cellcolor{cyan!20}0.32\% & \cellcolor{cyan!20}11.72\% \\
\midrule

\multirow{7}{*}{Code Generation}
& Aligned & 71.63\% & 0.00\% & 1.22\% & 51.96\% & 0.58\% & 0.00\% & 85.89\% & 0.38\% & 9.09\% \\
\cmidrule{2-11}
& SFT & 74.19\% & 2.25\% & 11.67\% & 52.63\% & 2.76\% & 5.76\% & 88.06\% & 0.64\% & 7.98\% \\
& \cellcolor{green!20}Weight Decay & \cellcolor{green!20}73.47\% & \cellcolor{green!20}1.67\% & \cellcolor{green!20}8.08\% & \cellcolor{green!20}\textbf{53.20}\% & \cellcolor{green!20}2.44\% & \cellcolor{green!20}6.97\% & \cellcolor{green!20}88.08\% & \cellcolor{green!20}0.71\% & \cellcolor{green!20}13.74\% \\
& \cellcolor{green!20}Dropout & \cellcolor{green!20}73.64\% & \cellcolor{green!20}1.73\% & \cellcolor{green!20}8.23\% & \cellcolor{green!20}53.17\% & \cellcolor{green!20}2.95\% & \cellcolor{green!20}5.96\% & \cellcolor{green!20}87.70\% & \cellcolor{green!20}0.83\% & \cellcolor{green!20}11.52\% \\
& \cellcolor{cyan!20}Linear & \cellcolor{cyan!20}\textbf{75.32}\% & \cellcolor{cyan!20}0.71\% & \cellcolor{cyan!20}\textbf{4.27}\% & \cellcolor{cyan!20}53.04\% & \cellcolor{cyan!20}1.73\% & \cellcolor{cyan!20}\textbf{3.03}\% & \cellcolor{cyan!20}89.37\% & \cellcolor{cyan!20}\textbf{0.32}\% & \cellcolor{cyan!20}7.88\% \\
& \cellcolor{cyan!20}DARE & \cellcolor{cyan!20}74.46\% & \cellcolor{cyan!20}\textbf{0.64}\% & \cellcolor{cyan!20}4.65\% & \cellcolor{cyan!20}53.09\% & \cellcolor{cyan!20}1.86\% & \cellcolor{cyan!20}3.74\% & \cellcolor{cyan!20}\textbf{89.64}\% & \cellcolor{cyan!20}0.51\% & \cellcolor{cyan!20}\textbf{7.07}\% \\
& \cellcolor{cyan!20}SLERP & \cellcolor{cyan!20}75.01\% & \cellcolor{cyan!20}0.71\% & \cellcolor{cyan!20}4.34\% & \cellcolor{cyan!20}53.07\% & \cellcolor{cyan!20}\textbf{1.67}\% & \cellcolor{cyan!20}3.23\% & \cellcolor{cyan!20}89.39\% & \cellcolor{cyan!20}\textbf{0.32}\% & \cellcolor{cyan!20}8.18\% \\

\midrule

\multirow{7}{*}{Tool Using Proficiency}
& Aligned & 0.8979\% & 0.00\% & 1.22\% & 0.7280 & 0.58\% & 0.00\% & 0.9357 & 0.38\% & 9.09\% \\
\cmidrule{2-11}
& SFT & 0.8989 & 0.83\% & 3.45\% & 0.8802 & \textbf{0.64}\% & \textbf{0.10}\% & 0.9369 & 0.58\% & 8.08\% \\
& \cellcolor{green!20}Weight Decay & \cellcolor{green!20}\textbf{0.9282} & \cellcolor{green!20}1.41\% & \cellcolor{green!20}3.22\% & \cellcolor{green!20}0.8838 & \cellcolor{green!20}0.77\% & \cellcolor{green!20}0.30\% & \cellcolor{green!20}0.9177 & \cellcolor{green!20}0.58\% & \cellcolor{green!20}\textbf{8.48}\% \\
& \cellcolor{green!20}Dropout & \cellcolor{green!20}0.9269 & \cellcolor{green!20}0.83\% & \cellcolor{green!20}1.92\% & \cellcolor{green!20}\textbf{0.8865} & \cellcolor{green!20}0.83\% & \cellcolor{green!20}0.40\% & \cellcolor{green!20}\textbf{0.9514} & \cellcolor{green!20}0.77\% & \cellcolor{green!20}10.90\% \\
& \cellcolor{cyan!20}Linear & \cellcolor{cyan!20}0.9266 & \cellcolor{cyan!20}0.77\% & \cellcolor{cyan!20}2.44\% & \cellcolor{cyan!20}0.8793 & \cellcolor{cyan!20}\textbf{0.64}\% & \cellcolor{cyan!20}0.20\% & \cellcolor{cyan!20}0.9489 & \cellcolor{cyan!20}0.13\% & \cellcolor{cyan!20}9.39\% \\
& \cellcolor{cyan!20}DARE & \cellcolor{cyan!20}0.9251 & \cellcolor{cyan!20}\textbf{0.45}\% & \cellcolor{cyan!20}\textbf{1.21}\% & \cellcolor{cyan!20}0.8793 & \cellcolor{cyan!20}\textbf{0.64}\% & \cellcolor{cyan!20}0.20\% & \cellcolor{cyan!20}0.9149 & \cellcolor{cyan!20}\textbf{0.06}\% & \cellcolor{cyan!20}9.39\% \\
& \cellcolor{cyan!20}SLERP & \cellcolor{cyan!20}0.9266 & \cellcolor{cyan!20}\textbf{0.44}\% & \cellcolor{cyan!20}1.72\% & \cellcolor{cyan!20}0.8802 & \cellcolor{cyan!20}\textbf{0.64}\% & \cellcolor{cyan!20}\textbf{0.10}\% & \cellcolor{cyan!20}0.9152 & \cellcolor{cyan!20}0.13\% & \cellcolor{cyan!20}9.19\% \\
\midrule

\end{tabular}
\end{adjustbox}
\label{tab:main}
\end{table*}

%% file: figures_tex/harm_cat.tex




\begin{figure*}[ht!]
    \centering

    \begin{minipage}[t]{0.45\linewidth}
        \centering
        \includegraphics[width=\linewidth]{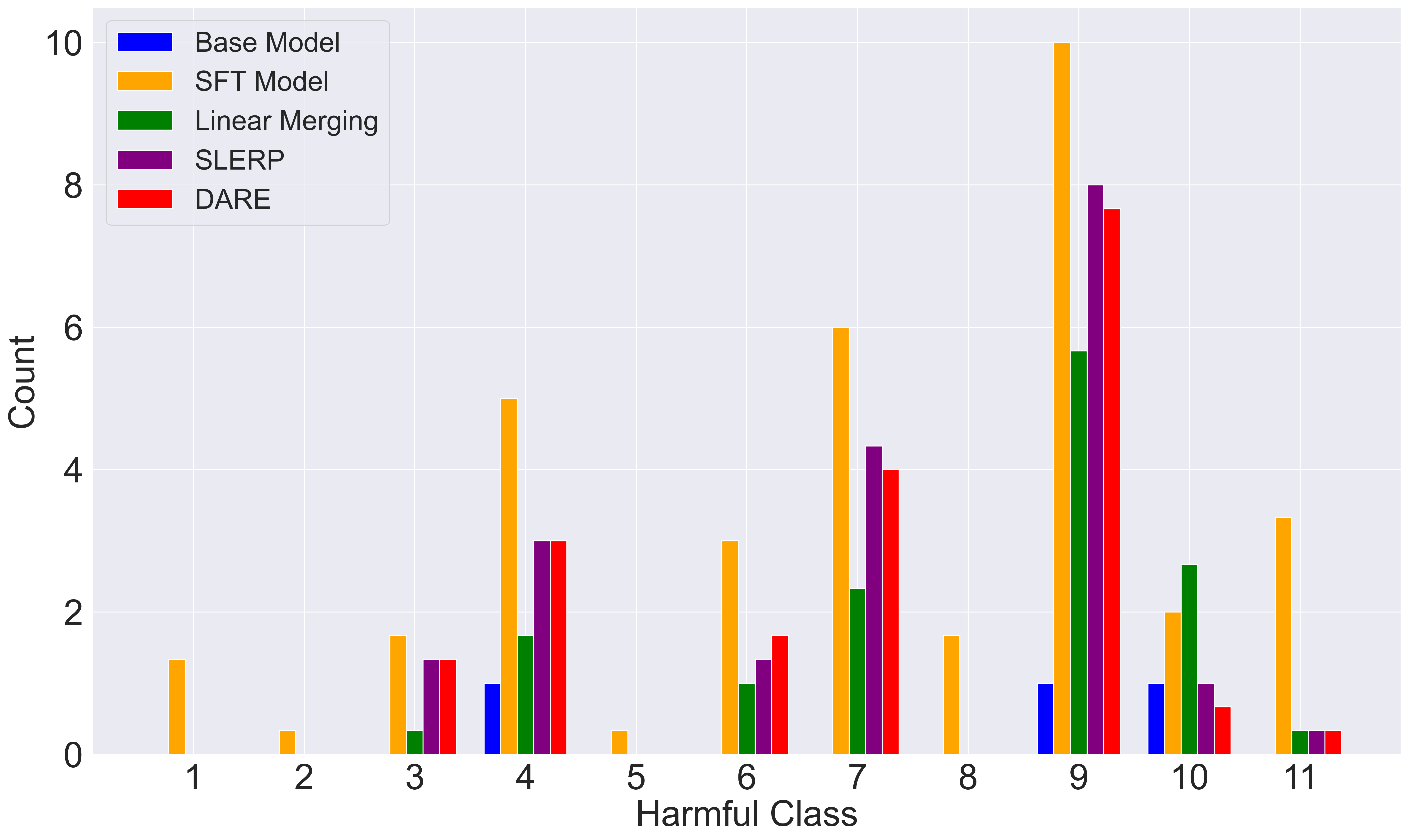}
        \footnotesize  LLaMA-3-8B-Instruct (Reasoning)
    \end{minipage}
    \hfill
    \begin{minipage}[t]{0.45\linewidth}
        \centering
        \includegraphics[width=\linewidth]{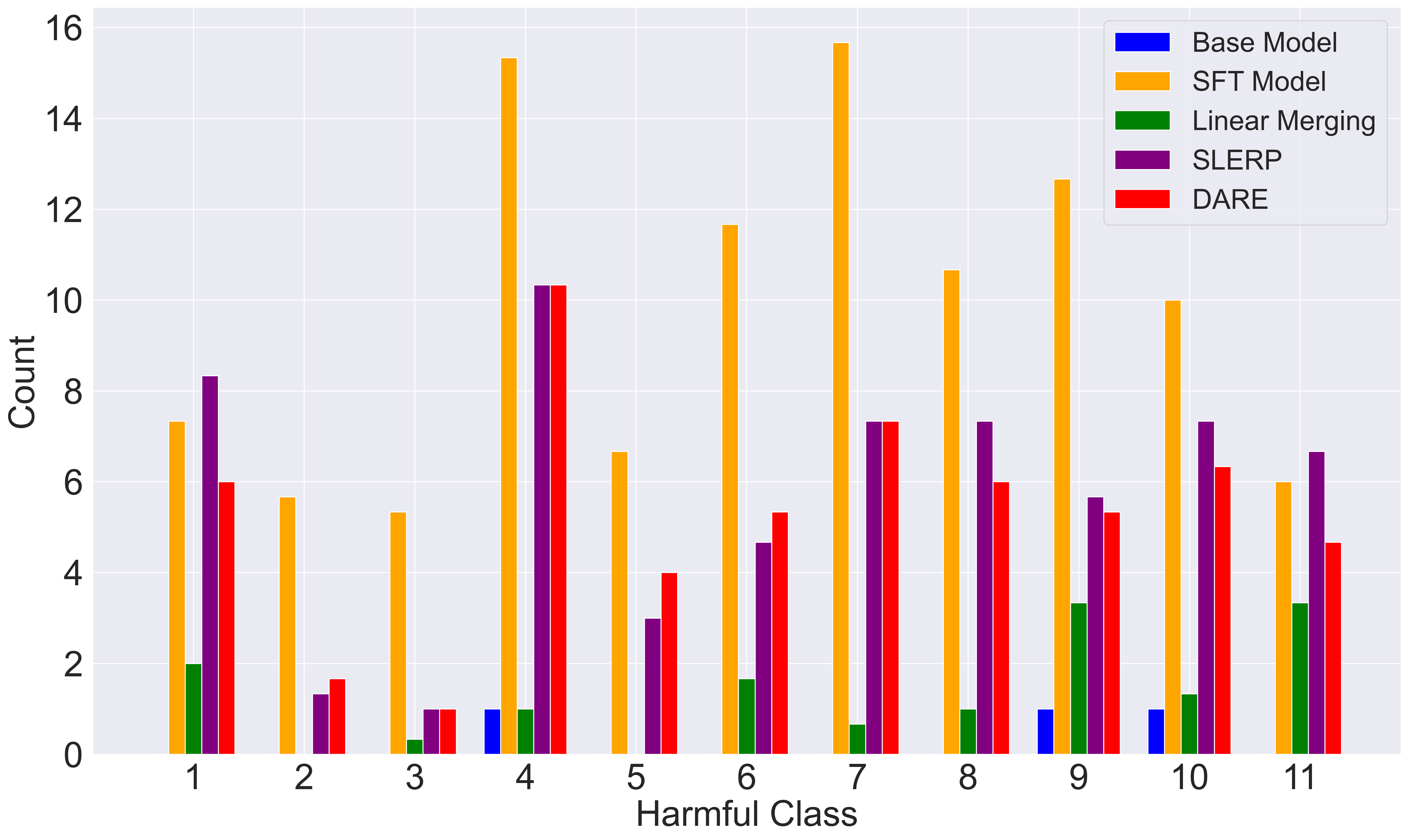}
        \footnotesize  LLaMA-3-8B-Instruct (Medical Assistance)
    \end{minipage}

    \vspace{2mm}

    \begin{minipage}[t]{0.45\linewidth}
        \centering
        \includegraphics[width=\linewidth]{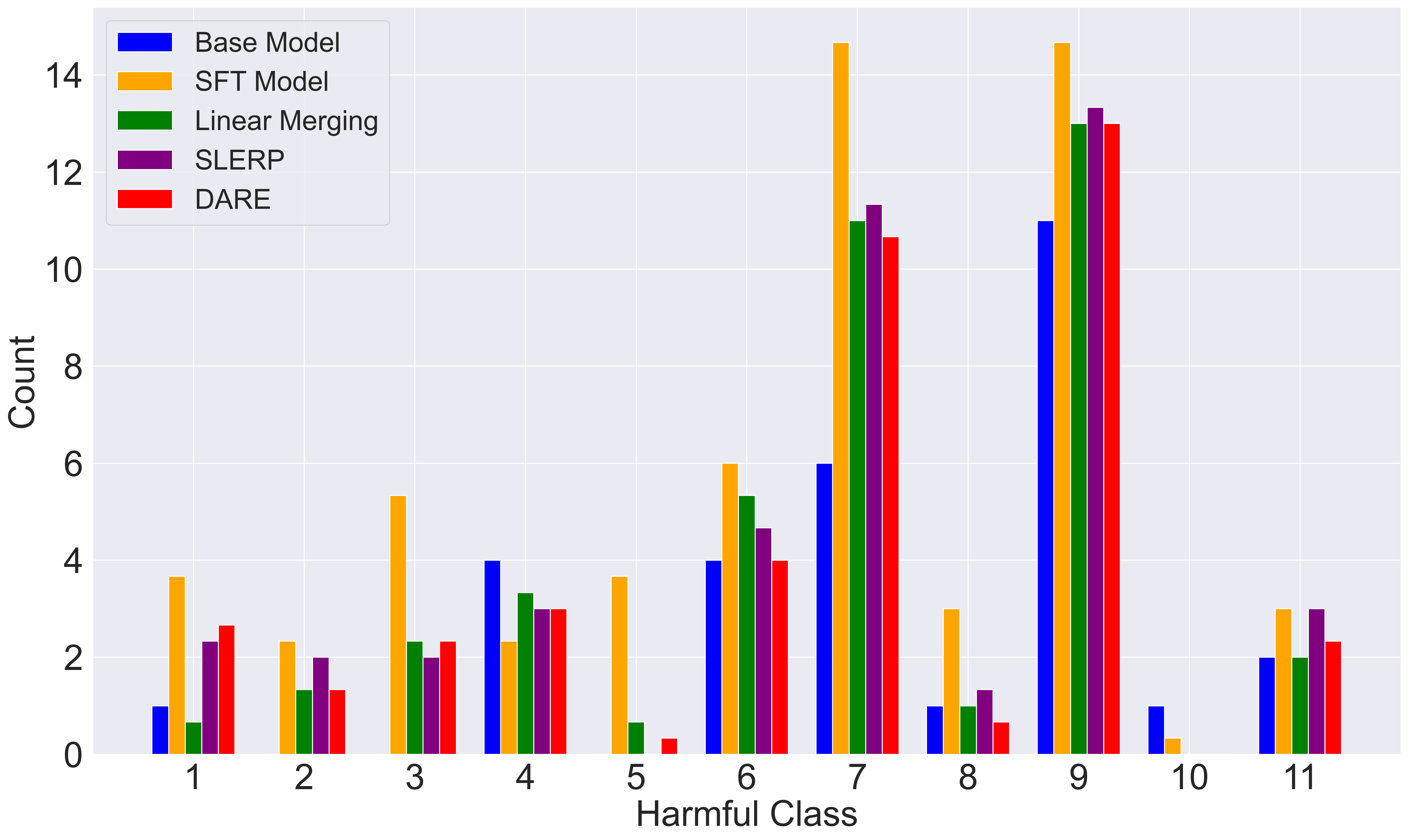}
        \footnotesize  Qwen2.5-7B-Instruct (Reasoning)
    \end{minipage}
    \hfill
    \begin{minipage}[t]{0.45\linewidth}
        \centering
        \includegraphics[width=\linewidth]{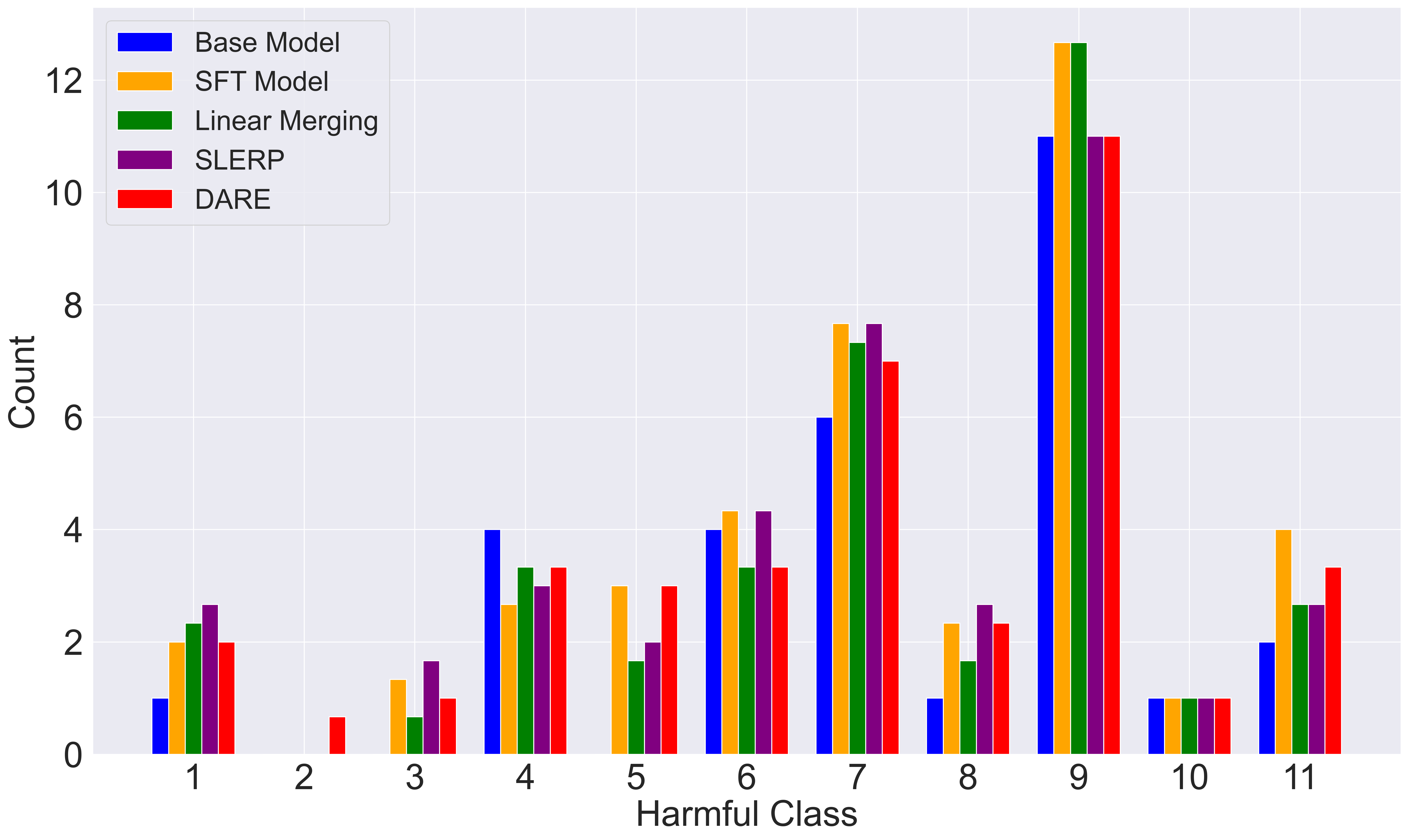}
        \footnotesize  Qwen2.5-7B-Instruct (Medical Assistance)
    \end{minipage}

    \caption{\textbf{Safety degradation across categories in the HEx-PHI benchmark.} ASR distributions over 11 harmful categories for LLaMA-3-8B-Instruct and Qwen2.5-7B-Instruct on the Reasoning and Medical Assistance tasks.}
    \label{fig:hexphi-category-breakdown}
\end{figure*}

%% file: main.bbl
\begin{thebibliography}{54}
\providecommand{\natexlab}[1]{#1}

\bibitem[{Bhardwaj et~al.(2024)Bhardwaj, Do, and Poria}]{homersimpson}
Rishabh Bhardwaj, Duc~Anh Do, and Soujanya Poria. 2024.
\newblock \href {https://doi.org/10.18653/v1/2024.acl-long.762} {Language models are {H}omer simpson! safety re-alignment of fine-tuned language models through task arithmetic}.
\newblock In \emph{Proceedings of the 62nd Annual Meeting of the Association for Computational Linguistics (Volume 1: Long Papers)}, pages 14138--14149, Bangkok, Thailand. Association for Computational Linguistics.

\bibitem[{Bianchi et~al.(2024)Bianchi, Suzgun, Attanasio, Rottger, Jurafsky, Hashimoto, and Zou}]{bianchi2024safetytuned}
Federico Bianchi, Mirac Suzgun, Giuseppe Attanasio, Paul Rottger, Dan Jurafsky, Tatsunori Hashimoto, and James Zou. 2024.
\newblock \href {https://openreview.net/forum?id=gT5hALch9z} {Safety-tuned {LL}a{MA}s: Lessons from improving the safety of large language models that follow instructions}.
\newblock In \emph{The Twelfth International Conference on Learning Representations}.

\bibitem[{Chen et~al.(2021)Chen, Tworek, Jun, Yuan, Pinto, Kaplan, Edwards, Burda, Joseph, Brockman et~al.}]{humaneval}
Mark Chen, Jerry Tworek, Heewoo Jun, Qiming Yuan, Henrique Ponde De~Oliveira Pinto, Jared Kaplan, Harri Edwards, Yuri Burda, Nicholas Joseph, Greg Brockman, and 1 others. 2021.
\newblock Evaluating large language models trained on code.
\newblock \emph{arXiv preprint arXiv:2107.03374}.

\bibitem[{Chen et~al.(2022)Chen, Gao, Cui, Qi, Huang, Liu, and Sun}]{advbench}
Yangyi Chen, Hongcheng Gao, Ganqu Cui, Fanchao Qi, Longtao Huang, Zhiyuan Liu, and Maosong Sun. 2022.
\newblock \href {https://doi.org/10.18653/v1/2022.emnlp-main.771} {Why should adversarial perturbations be imperceptible? rethink the research paradigm in adversarial {NLP}}.
\newblock In \emph{Proceedings of the 2022 Conference on Empirical Methods in Natural Language Processing}, pages 11222--11237, Abu Dhabi, United Arab Emirates. Association for Computational Linguistics.

\bibitem[{Chiang and Lee(2023)}]{chiang-lee-2023-large}
Cheng-Han Chiang and Hung-yi Lee. 2023.
\newblock \href {https://doi.org/10.18653/v1/2023.acl-long.870} {Can large language models be an alternative to human evaluations?}
\newblock In \emph{Proceedings of the 61st Annual Meeting of the Association for Computational Linguistics (Volume 1: Long Papers)}, pages 15607--15631, Toronto, Canada. Association for Computational Linguistics.

\bibitem[{Chung et~al.(2024)Chung, Hou, Longpre, Zoph, Tay, Fedus, Li, Wang, Dehghani, Brahma, Webson, Gu, Dai, Suzgun, Chen, Chowdhery, Castro-Ros, Pellat, Robinson, Valter, Narang, Mishra, Yu, Zhao, Huang, Dai, Yu, Petrov, Chi, Dean, Devlin, Roberts, Zhou, Le, and Wei}]{sft}
Hyung~Won Chung, Le~Hou, Shayne Longpre, Barret Zoph, Yi~Tay, William Fedus, Yunxuan Li, Xuezhi Wang, Mostafa Dehghani, Siddhartha Brahma, Albert Webson, Shixiang~Shane Gu, Zhuyun Dai, Mirac Suzgun, Xinyun Chen, Aakanksha Chowdhery, Alex Castro-Ros, Marie Pellat, Kevin Robinson, and 16 others. 2024.
\newblock \href {https://jmlr.org/papers/v25/23-0870.html} {Scaling instruction-finetuned language models}.
\newblock \emph{J. Mach. Learn. Res.}, 25:70:1--70:53.

\bibitem[{Gao et~al.(2024)Gao, Tow, Abbasi, Biderman, Black, DiPofi, Foster, Golding, Hsu, Le~Noac'h, Li, McDonell, Muennighoff, Ociepa, Phang, Reynolds, Schoelkopf, Skowron, Sutawika, Tang, Thite, Wang, Wang, and Zou}]{eval-harness}
Leo Gao, Jonathan Tow, Baber Abbasi, Stella Biderman, Sid Black, Anthony DiPofi, Charles Foster, Laurence Golding, Jeffrey Hsu, Alain Le~Noac'h, Haonan Li, Kyle McDonell, Niklas Muennighoff, Chris Ociepa, Jason Phang, Laria Reynolds, Hailey Schoelkopf, Aviya Skowron, Lintang Sutawika, and 5 others. 2024.
\newblock \href {https://doi.org/10.5281/zenodo.12608602} {A framework for few-shot language model evaluation}.

\bibitem[{Goddard et~al.(2024)Goddard, Siriwardhana, Ehghaghi, Meyers, Karpukhin, Benedict, McQuade, and Solawetz}]{mergekit}
Charles Goddard, Shamane Siriwardhana, Malikeh Ehghaghi, Luke Meyers, Vladimir Karpukhin, Brian Benedict, Mark McQuade, and Jacob Solawetz. 2024.
\newblock \href {https://doi.org/10.18653/v1/2024.emnlp-industry.36} {Arcee{'}s {M}erge{K}it: A toolkit for merging large language models}.
\newblock In \emph{Proceedings of the 2024 Conference on Empirical Methods in Natural Language Processing: Industry Track}, pages 477--485, Miami, Florida, US. Association for Computational Linguistics.

\bibitem[{Grattafiori et~al.(2024)Grattafiori, Dubey, Jauhri, Pandey, Kadian, Al-Dahle, Letman, Mathur, Schelten, Vaughan, Yang, Fan, Goyal, Hartshorn, Yang, Mitra, Sravankumar, Korenev, Hinsvark, Rao, Zhang, Rodriguez, Gregerson, Spataru, Roziere, Biron, Tang, Chern, Caucheteux, Nayak, Bi, Marra, McConnell, Keller, Touret, Wu, Wong, Ferrer, Nikolaidis, Allonsius, Song, Pintz, Livshits, Wyatt, Esiobu, Choudhary, Mahajan, Garcia-Olano, Perino, Hupkes, Lakomkin, AlBadawy, Lobanova, Dinan, Smith, Radenovic, Guzmán, Zhang, Synnaeve, Lee, Anderson, Thattai, Nail, Mialon, Pang, Cucurell, Nguyen, Korevaar, Xu, Touvron, Zarov, Ibarra, Kloumann, Misra, Evtimov, Zhang, Copet, Lee, Geffert, Vranes, Park, Mahadeokar, Shah, van~der Linde, Billock, Hong, Lee, Fu, Chi, Huang, Liu, Wang, Yu, Bitton, Spisak, Park, Rocca, Johnstun, Saxe, Jia, Alwala, Prasad, Upasani, Plawiak, Li, Heafield, Stone, El-Arini, Iyer, Malik, Chiu, Bhalla, Lakhotia, Rantala-Yeary, van~der Maaten, Chen, Tan, Jenkins, Martin, Madaan, Malo, Blecher,
  Landzaat, de~Oliveira, Muzzi, Pasupuleti, Singh, Paluri, Kardas, Tsimpoukelli, Oldham, Rita, Pavlova, Kambadur, Lewis, Si, Singh, Hassan, Goyal, Torabi, Bashlykov, Bogoychev, Chatterji, Zhang, Duchenne, Çelebi, Alrassy, Zhang, Li, Vasic, Weng, Bhargava, Dubal, Krishnan, Koura, Xu, He, Dong, Srinivasan, Ganapathy, Calderer, Cabral, Stojnic, Raileanu, Maheswari, Girdhar, Patel, Sauvestre, Polidoro, Sumbaly, Taylor, Silva, Hou, Wang, Hosseini, Chennabasappa, Singh, Bell, Kim, Edunov, Nie, Narang, Raparthy, Shen, Wan, Bhosale, Zhang, Vandenhende, Batra, Whitman, Sootla, Collot, Gururangan, Borodinsky, Herman, Fowler, Sheasha, Georgiou, Scialom, Speckbacher, Mihaylov, Xiao, Karn, Goswami, Gupta, Ramanathan, Kerkez, Gonguet, Do, Vogeti, Albiero, Petrovic, Chu, Xiong, Fu, Meers, Martinet, Wang, Wang, Tan, Xia, Xie, Jia, Wang, Goldschlag, Gaur, Babaei, Wen, Song, Zhang, Li, Mao, Coudert, Yan, Chen, Papakipos, Singh, Srivastava, Jain, Kelsey, Shajnfeld, Gangidi, Victoria, Goldstand, Menon, Sharma, Boesenberg,
  Baevski, Feinstein, Kallet, Sangani, Teo, Yunus, Lupu, Alvarado, Caples, Gu, Ho, Poulton, Ryan, Ramchandani, Dong, Franco, Goyal, Saraf, Chowdhury, Gabriel, Bharambe, Eisenman, Yazdan, James, Maurer, Leonhardi, Huang, Loyd, Paola, Paranjape, Liu, Wu, Ni, Hancock, Wasti, Spence, Stojkovic, Gamido, Montalvo, Parker, Burton, Mejia, Liu, Wang, Kim, Zhou, Hu, Chu, Cai, Tindal, Feichtenhofer, Gao, Civin, Beaty, Kreymer, Li, Adkins, Xu, Testuggine, David, Parikh, Liskovich, Foss, Wang, Le, Holland, Dowling, Jamil, Montgomery, Presani, Hahn, Wood, Le, Brinkman, Arcaute, Dunbar, Smothers, Sun, Kreuk, Tian, Kokkinos, Ozgenel, Caggioni, Kanayet, Seide, Florez, Schwarz, Badeer, Swee, Halpern, Herman, Sizov, Guangyi, Zhang, Lakshminarayanan, Inan, Shojanazeri, Zou, Wang, Zha, Habeeb, Rudolph, Suk, Aspegren, Goldman, Zhan, Damlaj, Molybog, Tufanov, Leontiadis, Veliche, Gat, Weissman, Geboski, Kohli, Lam, Asher, Gaya, Marcus, Tang, Chan, Zhen, Reizenstein, Teboul, Zhong, Jin, Yang, Cummings, Carvill, Shepard, McPhie,
  Torres, Ginsburg, Wang, Wu, U, Saxena, Khandelwal, Zand, Matosich, Veeraraghavan, Michelena, Li, Jagadeesh, Huang, Chawla, Huang, Chen, Garg, A, Silva, Bell, Zhang, Guo, Yu, Moshkovich, Wehrstedt, Khabsa, Avalani, Bhatt, Mankus, Hasson, Lennie, Reso, Groshev, Naumov, Lathi, Keneally, Liu, Seltzer, Valko, Restrepo, Patel, Vyatskov, Samvelyan, Clark, Macey, Wang, Hermoso, Metanat, Rastegari, Bansal, Santhanam, Parks, White, Bawa, Singhal, Egebo, Usunier, Mehta, Laptev, Dong, Cheng, Chernoguz, Hart, Salpekar, Kalinli, Kent, Parekh, Saab, Balaji, Rittner, Bontrager, Roux, Dollar, Zvyagina, Ratanchandani, Yuvraj, Liang, Alao, Rodriguez, Ayub, Murthy, Nayani, Mitra, Parthasarathy, Li, Hogan, Battey, Wang, Howes, Rinott, Mehta, Siby, Bondu, Datta, Chugh, Hunt, Dhillon, Sidorov, Pan, Mahajan, Verma, Yamamoto, Ramaswamy, Lindsay, Lindsay, Feng, Lin, Zha, Patil, Shankar, Zhang, Zhang, Wang, Agarwal, Sajuyigbe, Chintala, Max, Chen, Kehoe, Satterfield, Govindaprasad, Gupta, Deng, Cho, Virk, Subramanian, Choudhury,
  Goldman, Remez, Glaser, Best, Koehler, Robinson, Li, Zhang, Matthews, Chou, Shaked, Vontimitta, Ajayi, Montanez, Mohan, Kumar, Mangla, Ionescu, Poenaru, Mihailescu, Ivanov, Li, Wang, Jiang, Bouaziz, Constable, Tang, Wu, Wang, Wu, Gao, Kleinman, Chen, Hu, Jia, Qi, Li, Zhang, Zhang, Adi, Nam, Yu, Wang, Zhao, Hao, Qian, Li, He, Rait, DeVito, Rosnbrick, Wen, Yang, Zhao, and Ma}]{grattafiori2024llama3herdmodels}
Aaron Grattafiori, Abhimanyu Dubey, Abhinav Jauhri, Abhinav Pandey, Abhishek Kadian, Ahmad Al-Dahle, Aiesha Letman, Akhil Mathur, Alan Schelten, Alex Vaughan, Amy Yang, Angela Fan, Anirudh Goyal, Anthony Hartshorn, Aobo Yang, Archi Mitra, Archie Sravankumar, Artem Korenev, Arthur Hinsvark, and 542 others. 2024.
\newblock \href {https://arxiv.org/abs/2407.21783} {The llama 3 herd of models}.
\newblock \emph{Preprint}, arXiv:2407.21783.

\bibitem[{Han et~al.(2024)Han, Rao, Ettinger, Jiang, Lin, Lambert, Choi, and Dziri}]{han2024wildguard}
Seungju Han, Kavel Rao, Allyson Ettinger, Liwei Jiang, Bill~Yuchen Lin, Nathan Lambert, Yejin Choi, and Nouha Dziri. 2024.
\newblock \href {https://openreview.net/forum?id=Ich4tv4202} {Wildguard: Open one-stop moderation tools for safety risks, jailbreaks, and refusals of {LLM}s}.
\newblock In \emph{The Thirty-eight Conference on Neural Information Processing Systems Datasets and Benchmarks Track}.

\bibitem[{Hazra et~al.(2024)Hazra, Layek, Banerjee, and Poria}]{safetyarithmetic}
Rima Hazra, Sayan Layek, Somnath Banerjee, and Soujanya Poria. 2024.
\newblock \href {https://doi.org/10.18653/v1/2024.emnlp-main.1212} {Safety arithmetic: A framework for test-time safety alignment of language models by steering parameters and activations}.
\newblock In \emph{Proceedings of the 2024 Conference on Empirical Methods in Natural Language Processing}, pages 21759--21776, Miami, Florida, USA. Association for Computational Linguistics.

\bibitem[{Hsu et~al.(2024)Hsu, Tsai, Lin, Chen, Yu, and Huang}]{safelora}
Chia-Yi Hsu, Yu-Lin Tsai, Chih-Hsun Lin, Pin-Yu Chen, Chia-Mu Yu, and Chun-Ying Huang. 2024.
\newblock \href {https://openreview.net/forum?id=HcifdQZFZV} {Safe lo{RA}: The silver lining of reducing safety risks when finetuning large language models}.
\newblock In \emph{The Thirty-eighth Annual Conference on Neural Information Processing Systems}.

\bibitem[{Hu et~al.(2022)Hu, yelong shen, Wallis, Allen-Zhu, Li, Wang, Wang, and Chen}]{hu2022lora}
Edward~J Hu, yelong shen, Phillip Wallis, Zeyuan Allen-Zhu, Yuanzhi Li, Shean Wang, Lu~Wang, and Weizhu Chen. 2022.
\newblock \href {https://openreview.net/forum?id=nZeVKeeFYf9} {Lo{RA}: Low-rank adaptation of large language models}.
\newblock In \emph{International Conference on Learning Representations}.

\bibitem[{Huang et~al.(2024{\natexlab{a}})Huang, Li, Hsu, Chen, Lin, Hsiao, Tsai, and yi~Lee}]{chatvector}
Shih-Cheng Huang, Pin-Zu Li, Yu-Chi Hsu, Kuang-Ming Chen, Yu~Tung Lin, Shih-Kai Hsiao, Richard Tzong-Han Tsai, and Hung yi~Lee. 2024{\natexlab{a}}.
\newblock \href {https://arxiv.org/abs/2310.04799} {Chat vector: A simple approach to equip llms with instruction following and model alignment in new languages}.
\newblock \emph{Preprint}, arXiv:2310.04799.

\bibitem[{Huang et~al.(2024{\natexlab{b}})Huang, Bhattacharya, Joshi, Kimball, and Liu}]{antidote}
Tiansheng Huang, Gautam Bhattacharya, Pratik Joshi, Josh Kimball, and Ling Liu. 2024{\natexlab{b}}.
\newblock \href {https://arxiv.org/abs/2408.09600} {Antidote: Post-fine-tuning safety alignment for large language models against harmful fine-tuning}.
\newblock \emph{Preprint}, arXiv:2408.09600.

\bibitem[{Huang et~al.(2024{\natexlab{c}})Huang, Hu, Ilhan, Tekin, and Liu}]{huang2024lisa}
Tiansheng Huang, Sihao Hu, Fatih Ilhan, Selim~Furkan Tekin, and Ling Liu. 2024{\natexlab{c}}.
\newblock \href {https://openreview.net/forum?id=RPChapuXlC} {Lisa: Lazy safety alignment for large language models against harmful fine-tuning attack}.
\newblock In \emph{The Thirty-eighth Annual Conference on Neural Information Processing Systems}.

\bibitem[{Huang et~al.(2024{\natexlab{d}})Huang, Hu, and Liu}]{huang2024vaccine}
Tiansheng Huang, Sihao Hu, and Ling Liu. 2024{\natexlab{d}}.
\newblock \href {https://openreview.net/forum?id=lpXDZKiAnt} {Vaccine: Perturbation-aware alignment for large language models against harmful fine-tuning attack}.
\newblock In \emph{The Thirty-eighth Annual Conference on Neural Information Processing Systems}.

\bibitem[{Huang et~al.(2023)Huang, Ruan, Huang, Jin, Dong, Wu, Bensalem, Mu, Qi, Zhao, Cai, Zhang, Wu, Xu, Wu, Freitas, and Mustafa}]{huang2023surveysafetytrustworthinesslarge}
Xiaowei Huang, Wenjie Ruan, Wei Huang, Gaojie Jin, Yi~Dong, Changshun Wu, Saddek Bensalem, Ronghui Mu, Yi~Qi, Xingyu Zhao, Kaiwen Cai, Yanghao Zhang, Sihao Wu, Peipei Xu, Dengyu Wu, Andre Freitas, and Mustafa~A. Mustafa. 2023.
\newblock \href {https://arxiv.org/abs/2305.11391} {A survey of safety and trustworthiness of large language models through the lens of verification and validation}.
\newblock \emph{Preprint}, arXiv:2305.11391.

\bibitem[{Ilharco et~al.(2023)Ilharco, Ribeiro, Wortsman, Schmidt, Hajishirzi, and Farhadi}]{taskvector}
Gabriel Ilharco, Marco~Tulio Ribeiro, Mitchell Wortsman, Ludwig Schmidt, Hannaneh Hajishirzi, and Ali Farhadi. 2023.
\newblock \href {https://openreview.net/forum?id=6t0Kwf8-jrj} {Editing models with task arithmetic}.
\newblock In \emph{The Eleventh International Conference on Learning Representations}.

\bibitem[{Kirkpatrick et~al.(2017)Kirkpatrick, Pascanu, Rabinowitz, Veness, Desjardins, Rusu, Milan, Quan, Ramalho, Grabska-Barwinska, Hassabis, Clopath, Kumaran, and Hadsell}]{overcomingcatastrophic}
James Kirkpatrick, Razvan Pascanu, Neil Rabinowitz, Joel Veness, Guillaume Desjardins, Andrei~A. Rusu, Kieran Milan, John Quan, Tiago Ramalho, Agnieszka Grabska-Barwinska, Demis Hassabis, Claudia Clopath, Dharshan Kumaran, and Raia Hadsell. 2017.
\newblock \href {https://doi.org/10.1073/pnas.1611835114} {Overcoming catastrophic forgetting in neural networks}.
\newblock \emph{Proceedings of the National Academy of Sciences}, 114(13):3521–3526.

\bibitem[{Kwon et~al.(2023)Kwon, Li, Zhuang, Sheng, Zheng, Yu, Gonzalez, Zhang, and Stoica}]{vllm}
Woosuk Kwon, Zhuohan Li, Siyuan Zhuang, Ying Sheng, Lianmin Zheng, Cody~Hao Yu, Joseph~E. Gonzalez, Hao Zhang, and Ion Stoica. 2023.
\newblock Efficient memory management for large language model serving with pagedattention.
\newblock In \emph{Proceedings of the ACM SIGOPS 29th Symposium on Operating Systems Principles}.

\bibitem[{Li and Lee(2024)}]{li2024examiningforgettingcontinualpretraining}
Chen-An Li and Hung-Yi Lee. 2024.
\newblock \href {https://arxiv.org/abs/2401.03129} {Examining forgetting in continual pre-training of aligned large language models}.
\newblock \emph{Preprint}, arXiv:2401.03129.

\bibitem[{Li et~al.(2023)Li, Li, Zhang, Dan, Jiang, and Zhang}]{li2023chatdoctormedicalchatmodel}
Yunxiang Li, Zihan Li, Kai Zhang, Ruilong Dan, Steve Jiang, and You Zhang. 2023.
\newblock \href {https://arxiv.org/abs/2303.14070} {Chatdoctor: A medical chat model fine-tuned on a large language model meta-ai (llama) using medical domain knowledge}.
\newblock \emph{Preprint}, arXiv:2303.14070.

\bibitem[{Liu et~al.(2023)Liu, Iter, Xu, Wang, Xu, and Zhu}]{geval}
Yang Liu, Dan Iter, Yichong Xu, Shuohang Wang, Ruochen Xu, and Chenguang Zhu. 2023.
\newblock \href {https://doi.org/10.18653/v1/2023.emnlp-main.153} {{G}-eval: {NLG} evaluation using gpt-4 with better human alignment}.
\newblock In \emph{Proceedings of the 2023 Conference on Empirical Methods in Natural Language Processing}, pages 2511--2522, Singapore. Association for Computational Linguistics.

\bibitem[{Longpre et~al.(2023)Longpre, Hou, Vu, Webson, Chung, Tay, Zhou, Le, Zoph, Wei, and Roberts}]{longpre2023flancollectiondesigningdata}
Shayne Longpre, Le~Hou, Tu~Vu, Albert Webson, Hyung~Won Chung, Yi~Tay, Denny Zhou, Quoc~V. Le, Barret Zoph, Jason Wei, and Adam Roberts. 2023.
\newblock \href {https://arxiv.org/abs/2301.13688} {The flan collection: Designing data and methods for effective instruction tuning}.
\newblock \emph{Preprint}, arXiv:2301.13688.

\bibitem[{Loshchilov and Hutter(2019)}]{weightdecay}
Ilya Loshchilov and Frank Hutter. 2019.
\newblock \href {https://openreview.net/forum?id=Bkg6RiCqY7} {Decoupled weight decay regularization}.
\newblock In \emph{International Conference on Learning Representations}.

\bibitem[{Luo et~al.(2025)Luo, Yang, Meng, Li, Zhou, and Zhang}]{luo2025empiricalstudycatastrophicforgetting}
Yun Luo, Zhen Yang, Fandong Meng, Yafu Li, Jie Zhou, and Yue Zhang. 2025.
\newblock \href {https://arxiv.org/abs/2308.08747} {An empirical study of catastrophic forgetting in large language models during continual fine-tuning}.
\newblock \emph{Preprint}, arXiv:2308.08747.

\bibitem[{O'Brien et~al.(2024)O'Brien, Majercak, Fernandes, Edgar, Chen, Nori, Carignan, Horvitz, and Poursabzi-Sangde}]{obrien2024steeringlanguagemodelrefusal}
Kyle O'Brien, David Majercak, Xavier Fernandes, Richard Edgar, Jingya Chen, Harsha Nori, Dean Carignan, Eric Horvitz, and Forough Poursabzi-Sangde. 2024.
\newblock \href {https://arxiv.org/abs/2411.11296} {Steering language model refusal with sparse autoencoders}.
\newblock \emph{Preprint}, arXiv:2411.11296.

\bibitem[{OpenAI et~al.(2024)OpenAI, Achiam, Adler, Agarwal, Ahmad, Akkaya, Aleman, Almeida, Altenschmidt, Altman, Anadkat, Avila, Babuschkin, Balaji, Balcom, Baltescu, Bao, Bavarian, Belgum, Bello, Berdine, Bernadett-Shapiro, Berner, Bogdonoff, Boiko, Boyd, Brakman, Brockman, Brooks, Brundage, Button, Cai, Campbell, Cann, Carey, Carlson, Carmichael, Chan, Chang, Chantzis, Chen, Chen, Chen, Chen, Chen, Chess, Cho, Chu, Chung, Cummings, Currier, Dai, Decareaux, Degry, Deutsch, Deville, Dhar, Dohan, Dowling, Dunning, Ecoffet, Eleti, Eloundou, Farhi, Fedus, Felix, Fishman, Forte, Fulford, Gao, Georges, Gibson, Goel, Gogineni, Goh, Gontijo-Lopes, Gordon, Grafstein, Gray, Greene, Gross, Gu, Guo, Hallacy, Han, Harris, He, Heaton, Heidecke, Hesse, Hickey, Hickey, Hoeschele, Houghton, Hsu, Hu, Hu, Huizinga, Jain, Jain, Jang, Jiang, Jiang, Jin, Jin, Jomoto, Jonn, Jun, Kaftan, Łukasz Kaiser, Kamali, Kanitscheider, Keskar, Khan, Kilpatrick, Kim, Kim, Kim, Kirchner, Kiros, Knight, Kokotajlo, Łukasz Kondraciuk,
  Kondrich, Konstantinidis, Kosic, Krueger, Kuo, Lampe, Lan, Lee, Leike, Leung, Levy, Li, Lim, Lin, Lin, Litwin, Lopez, Lowe, Lue, Makanju, Malfacini, Manning, Markov, Markovski, Martin, Mayer, Mayne, McGrew, McKinney, McLeavey, McMillan, McNeil, Medina, Mehta, Menick, Metz, Mishchenko, Mishkin, Monaco, Morikawa, Mossing, Mu, Murati, Murk, Mély, Nair, Nakano, Nayak, Neelakantan, Ngo, Noh, Ouyang, O'Keefe, Pachocki, Paino, Palermo, Pantuliano, Parascandolo, Parish, Parparita, Passos, Pavlov, Peng, Perelman, de~Avila Belbute~Peres, Petrov, de~Oliveira~Pinto, Michael, Pokorny, Pokrass, Pong, Powell, Power, Power, Proehl, Puri, Radford, Rae, Ramesh, Raymond, Real, Rimbach, Ross, Rotsted, Roussez, Ryder, Saltarelli, Sanders, Santurkar, Sastry, Schmidt, Schnurr, Schulman, Selsam, Sheppard, Sherbakov, Shieh, Shoker, Shyam, Sidor, Sigler, Simens, Sitkin, Slama, Sohl, Sokolowsky, Song, Staudacher, Such, Summers, Sutskever, Tang, Tezak, Thompson, Tillet, Tootoonchian, Tseng, Tuggle, Turley, Tworek, Uribe, Vallone,
  Vijayvergiya, Voss, Wainwright, Wang, Wang, Wang, Ward, Wei, Weinmann, Welihinda, Welinder, Weng, Weng, Wiethoff, Willner, Winter, Wolrich, Wong, Workman, Wu, Wu, Wu, Xiao, Xu, Yoo, Yu, Yuan, Zaremba, Zellers, Zhang, Zhang, Zhao, Zheng, Zhuang, Zhuk, and Zoph}]{openai2024gpt4technicalreport}
OpenAI, Josh Achiam, Steven Adler, Sandhini Agarwal, Lama Ahmad, Ilge Akkaya, Florencia~Leoni Aleman, Diogo Almeida, Janko Altenschmidt, Sam Altman, Shyamal Anadkat, Red Avila, Igor Babuschkin, Suchir Balaji, Valerie Balcom, Paul Baltescu, Haiming Bao, Mohammad Bavarian, Jeff Belgum, and 262 others. 2024.
\newblock \href {https://arxiv.org/abs/2303.08774} {Gpt-4 technical report}.
\newblock \emph{Preprint}, arXiv:2303.08774.

\bibitem[{Ouyang et~al.(2022)Ouyang, Wu, Jiang, Almeida, Wainwright, Mishkin, Zhang, Agarwal, Slama, Ray, Schulman, Hilton, Kelton, Miller, Simens, Askell, Welinder, Christiano, Leike, and Lowe}]{ouyang2022traininglanguagemodelsfollow}
Long Ouyang, Jeff Wu, Xu~Jiang, Diogo Almeida, Carroll~L. Wainwright, Pamela Mishkin, Chong Zhang, Sandhini Agarwal, Katarina Slama, Alex Ray, John Schulman, Jacob Hilton, Fraser Kelton, Luke Miller, Maddie Simens, Amanda Askell, Peter Welinder, Paul Christiano, Jan Leike, and Ryan Lowe. 2022.
\newblock \href {https://arxiv.org/abs/2203.02155} {Training language models to follow instructions with human feedback}.
\newblock \emph{Preprint}, arXiv:2203.02155.

\bibitem[{Patil et~al.(2023)Patil, Zhang, Wang, and Gonzalez}]{openfunction}
Shishir~G. Patil, Tianjun Zhang, Xin Wang, and Joseph~E. Gonzalez. 2023.
\newblock Gorilla: Large language model connected with massive apis.

\bibitem[{Qi et~al.(2024)Qi, Zeng, Xie, Chen, Jia, Mittal, and Henderson}]{qi2024finetuning}
Xiangyu Qi, Yi~Zeng, Tinghao Xie, Pin-Yu Chen, Ruoxi Jia, Prateek Mittal, and Peter Henderson. 2024.
\newblock \href {https://openreview.net/forum?id=hTEGyKf0dZ} {Fine-tuning aligned language models compromises safety, even when users do not intend to!}
\newblock In \emph{The Twelfth International Conference on Learning Representations}.

\bibitem[{Rafailov et~al.(2023)Rafailov, Sharma, Mitchell, Manning, Ermon, and Finn}]{dpo}
Rafael Rafailov, Archit Sharma, Eric Mitchell, Christopher~D Manning, Stefano Ermon, and Chelsea Finn. 2023.
\newblock \href {https://openreview.net/forum?id=HPuSIXJaa9} {Direct preference optimization: Your language model is secretly a reward model}.
\newblock In \emph{Thirty-seventh Conference on Neural Information Processing Systems}.

\bibitem[{Srivastava et~al.(2014)Srivastava, Hinton, Krizhevsky, Sutskever, and Salakhutdinov}]{dropout}
Nitish Srivastava, Geoffrey Hinton, Alex Krizhevsky, Ilya Sutskever, and Ruslan Salakhutdinov. 2014.
\newblock \href {http://jmlr.org/papers/v15/srivastava14a.html} {Dropout: A simple way to prevent neural networks from overfitting}.
\newblock \emph{Journal of Machine Learning Research}, 15(56):1929--1958.

\bibitem[{Su et~al.(2024)Su, Farn, Sun, Chen, and Lee}]{syn2real}
Hsuan Su, Hua Farn, Fan-Yun Sun, Shang-Tse Chen, and Hung-yi Lee. 2024.
\newblock \href {https://doi.org/10.18653/v1/2024.emnlp-main.503} {Task arithmetic can mitigate synthetic-to-real gap in automatic speech recognition}.
\newblock In \emph{Proceedings of the 2024 Conference on Empirical Methods in Natural Language Processing}, pages 8905--8915, Miami, Florida, USA. Association for Computational Linguistics.

\bibitem[{Suzgun et~al.(2023)Suzgun, Scales, Sch{\"a}rli, Gehrmann, Tay, Chung, Chowdhery, Le, Chi, Zhou, and Wei}]{BBH}
Mirac Suzgun, Nathan Scales, Nathanael Sch{\"a}rli, Sebastian Gehrmann, Yi~Tay, Hyung~Won Chung, Aakanksha Chowdhery, Quoc Le, Ed~Chi, Denny Zhou, and Jason Wei. 2023.
\newblock \href {https://doi.org/10.18653/v1/2023.findings-acl.824} {Challenging {BIG}-bench tasks and whether chain-of-thought can solve them}.
\newblock In \emph{Findings of the Association for Computational Linguistics: ACL 2023}, pages 13003--13051, Toronto, Canada. Association for Computational Linguistics.

\bibitem[{Team et~al.(2024)Team, Mesnard, Hardin, Dadashi, Bhupatiraju, Pathak, Sifre, Rivière, Kale, Love, Tafti, Hussenot, Sessa, Chowdhery, Roberts, Barua, Botev, Castro-Ros, Slone, Héliou, Tacchetti, Bulanova, Paterson, Tsai, Shahriari, Lan, Choquette-Choo, Crepy, Cer, Ippolito, Reid, Buchatskaya, Ni, Noland, Yan, Tucker, Muraru, Rozhdestvenskiy, Michalewski, Tenney, Grishchenko, Austin, Keeling, Labanowski, Lespiau, Stanway, Brennan, Chen, Ferret, Chiu, Mao-Jones, Lee, Yu, Millican, Sjoesund, Lee, Dixon, Reid, Mikuła, Wirth, Sharman, Chinaev, Thain, Bachem, Chang, Wahltinez, Bailey, Michel, Yotov, Chaabouni, Comanescu, Jana, Anil, McIlroy, Liu, Mullins, Smith, Borgeaud, Girgin, Douglas, Pandya, Shakeri, De, Klimenko, Hennigan, Feinberg, Stokowiec, hui Chen, Ahmed, Gong, Warkentin, Peran, Giang, Farabet, Vinyals, Dean, Kavukcuoglu, Hassabis, Ghahramani, Eck, Barral, Pereira, Collins, Joulin, Fiedel, Senter, Andreev, and Kenealy}]{gemmateam2024gemmaopenmodelsbased}
Gemma Team, Thomas Mesnard, Cassidy Hardin, Robert Dadashi, Surya Bhupatiraju, Shreya Pathak, Laurent Sifre, Morgane Rivière, Mihir~Sanjay Kale, Juliette Love, Pouya Tafti, Léonard Hussenot, Pier~Giuseppe Sessa, Aakanksha Chowdhery, Adam Roberts, Aditya Barua, Alex Botev, Alex Castro-Ros, Ambrose Slone, and 89 others. 2024.
\newblock \href {https://arxiv.org/abs/2403.08295} {Gemma: Open models based on gemini research and technology}.
\newblock \emph{Preprint}, arXiv:2403.08295.

\bibitem[{Team(2024)}]{qwen2.5}
Qwen Team. 2024.
\newblock \href {https://qwenlm.github.io/blog/qwen2.5/} {Qwen2.5: A party of foundation models}.

\bibitem[{Wei et~al.(2024)Wei, Wang, Liu, Ding, and Zhang}]{wei2024magicoder}
Yuxiang Wei, Zhe Wang, Jiawei Liu, Yifeng Ding, and Lingming Zhang. 2024.
\newblock \href {https://proceedings.mlr.press/v235/wei24h.html} {Magicoder: Empowering code generation with {OSS}-instruct}.
\newblock In \emph{Proceedings of the 41st International Conference on Machine Learning}, volume 235 of \emph{Proceedings of Machine Learning Research}, pages 52632--52657. PMLR.

\bibitem[{White(2017)}]{slerp}
Tom White. 2017.
\newblock \href {https://openreview.net/forum?id=SypU81Ole} {Sampling generative networks}.

\bibitem[{Wortsman et~al.(2022{\natexlab{a}})Wortsman, Ilharco, Gadre, Roelofs, Gontijo-Lopes, Morcos, Namkoong, Farhadi, Carmon, Kornblith, and Schmidt}]{linearmerging}
Mitchell Wortsman, Gabriel Ilharco, Samir~Ya Gadre, Rebecca Roelofs, Raphael Gontijo-Lopes, Ari~S Morcos, Hongseok Namkoong, Ali Farhadi, Yair Carmon, Simon Kornblith, and Ludwig Schmidt. 2022{\natexlab{a}}.
\newblock \href {https://proceedings.mlr.press/v162/wortsman22a.html} {Model soups: averaging weights of multiple fine-tuned models improves accuracy without increasing inference time}.
\newblock In \emph{Proceedings of the 39th International Conference on Machine Learning}, volume 162 of \emph{Proceedings of Machine Learning Research}, pages 23965--23998. PMLR.

\bibitem[{Wortsman et~al.(2022{\natexlab{b}})Wortsman, Ilharco, Kim, Li, Hajishirzi, Farhadi, Namkoong, and Schmidt}]{wortsman2022robust}
Mitchell Wortsman, Gabriel Ilharco, Jong~Wook Kim, Mike Li, Hanna Hajishirzi, Ali Farhadi, Hongseok Namkoong, and Ludwig Schmidt. 2022{\natexlab{b}}.
\newblock \href {https://openreview.net/forum?id=yrbF6ekqQ9w} {Robust fine-tuning of zero-shot models}.

\bibitem[{Wu et~al.(2025)Wu, Lu, Zhao, and Qin}]{seperatethewheat}
Di~Wu, Xin Lu, Yanyan Zhao, and Bing Qin. 2025.
\newblock \href {https://doi.org/10.18653/v1/2025.findings-acl.66} {Separate the wheat from the chaff: A post-hoc approach to safety re-alignment for fine-tuned language models}.
\newblock In \emph{Findings of the Association for Computational Linguistics: ACL 2025}, pages 1210--1225, Vienna, Austria. Association for Computational Linguistics.

\bibitem[{Xie et~al.(2025)Xie, Qi, Zeng, Huang, Sehwag, Huang, He, Wei, Li, Sheng, Jia, Li, Li, Chen, Henderson, and Mittal}]{xie2025sorrybench}
Tinghao Xie, Xiangyu Qi, Yi~Zeng, Yangsibo Huang, Udari~Madhushani Sehwag, Kaixuan Huang, Luxi He, Boyi Wei, Dacheng Li, Ying Sheng, Ruoxi Jia, Bo~Li, Kai Li, Danqi Chen, Peter Henderson, and Prateek Mittal. 2025.
\newblock \href {https://openreview.net/forum?id=YfKNaRktan} {{SORRY}-bench: Systematically evaluating large language model safety refusal}.
\newblock In \emph{The Thirteenth International Conference on Learning Representations}.

\bibitem[{Xu et~al.(2024)Xu, Liu, Deng, Li, and Picek}]{jailbreaksurvey}
Zihao Xu, Yi~Liu, Gelei Deng, Yuekang Li, and Stjepan Picek. 2024.
\newblock \href {https://doi.org/10.18653/v1/2024.findings-acl.443} {A comprehensive study of jailbreak attack versus defense for large language models}.
\newblock In \emph{Findings of the Association for Computational Linguistics: ACL 2024}, pages 7432--7449, Bangkok, Thailand. Association for Computational Linguistics.

\bibitem[{Yang et~al.(2023)Yang, Wang, Zhang, Petzold, Wang, Zhao, and Lin}]{yang2023shadowalignmenteasesubverting}
Xianjun Yang, Xiao Wang, Qi~Zhang, Linda Petzold, William~Yang Wang, Xun Zhao, and Dahua Lin. 2023.
\newblock \href {https://arxiv.org/abs/2310.02949} {Shadow alignment: The ease of subverting safely-aligned language models}.
\newblock \emph{Preprint}, arXiv:2310.02949.

\bibitem[{Yang et~al.(2024)Yang, Pang, Feng, Wang, Chen, Zhu, and Liu}]{selfdistill}
Zhaorui Yang, Tianyu Pang, Haozhe Feng, Han Wang, Wei Chen, Minfeng Zhu, and Qian Liu. 2024.
\newblock \href {https://doi.org/10.18653/v1/2024.acl-long.58} {Self-distillation bridges distribution gap in language model fine-tuning}.
\newblock In \emph{Proceedings of the 62nd Annual Meeting of the Association for Computational Linguistics (Volume 1: Long Papers)}, pages 1028--1043, Bangkok, Thailand. Association for Computational Linguistics.

\bibitem[{Yi et~al.(2024)Yi, Zheng, Wang, Wang, and He}]{yi2024safetyrealignmentframeworksubspaceoriented}
Xin Yi, Shunfan Zheng, Linlin Wang, Xiaoling Wang, and Liang He. 2024.
\newblock \href {https://arxiv.org/abs/2405.09055} {A safety realignment framework via subspace-oriented model fusion for large language models}.
\newblock \emph{Preprint}, arXiv:2405.09055.

\bibitem[{Yu et~al.(2024)Yu, Yu, Yu, Huang, and Li}]{dare}
Le~Yu, Bowen Yu, Haiyang Yu, Fei Huang, and Yongbin Li. 2024.
\newblock \href {https://openreview.net/forum?id=fq0NaiU8Ex} {Language models are super mario: Absorbing abilities from homologous models as a free lunch}.
\newblock In \emph{Forty-first International Conference on Machine Learning}.

\bibitem[{Zhan et~al.(2024)Zhan, Fang, Bindu, Gupta, Hashimoto, and Kang}]{zhan-etal-2024-removing}
Qiusi Zhan, Richard Fang, Rohan Bindu, Akul Gupta, Tatsunori Hashimoto, and Daniel Kang. 2024.
\newblock \href {https://doi.org/10.18653/v1/2024.naacl-short.59} {Removing {RLHF} protections in {GPT}-4 via fine-tuning}.
\newblock In \emph{Proceedings of the 2024 Conference of the North American Chapter of the Association for Computational Linguistics: Human Language Technologies (Volume 2: Short Papers)}, pages 681--687, Mexico City, Mexico. Association for Computational Linguistics.

\bibitem[{Zhang* et~al.(2020)Zhang*, Kishore*, Wu*, Weinberger, and Artzi}]{Zhang*2020BERTScore:}
Tianyi Zhang*, Varsha Kishore*, Felix Wu*, Kilian~Q. Weinberger, and Yoav Artzi. 2020.
\newblock \href {https://openreview.net/forum?id=SkeHuCVFDr} {Bertscore: Evaluating text generation with bert}.
\newblock In \emph{International Conference on Learning Representations}.

\bibitem[{Zheng et~al.(2024)Zheng, Zhang, Zhang, Ye, Luo, Feng, and Ma}]{zheng2024llamafactory}
Yaowei Zheng, Richong Zhang, Junhao Zhang, Yanhan Ye, Zheyan Luo, Zhangchi Feng, and Yongqiang Ma. 2024.
\newblock \href {http://arxiv.org/abs/2403.13372} {Llamafactory: Unified efficient fine-tuning of 100+ language models}.
\newblock In \emph{Proceedings of the 62nd Annual Meeting of the Association for Computational Linguistics (Volume 3: System Demonstrations)}, Bangkok, Thailand. Association for Computational Linguistics.

\bibitem[{Zhou et~al.(2023)Zhou, Lu, Mishra, Brahma, Basu, Luan, Zhou, and Hou}]{ifeval}
Jeffrey Zhou, Tianjian Lu, Swaroop Mishra, Siddhartha Brahma, Sujoy Basu, Yi~Luan, Denny Zhou, and Le~Hou. 2023.
\newblock \href {https://arxiv.org/abs/2311.07911} {Instruction-following evaluation for large language models}.
\newblock \emph{Preprint}, arXiv:2311.07911.

\bibitem[{Zong et~al.(2024)Zong, Bohdal, Yu, Yang, and Timothy}]{safetyfinetuningatnocost}
Yongshuo Zong, Ondrej Bohdal, Tingyang Yu, Yongxin Yang, and Hospedales Timothy. 2024.
\newblock Safety fine-tuning at (almost) no cost: A baseline for vision large language models.
\newblock \emph{arXiv preprint arXiv:2402.02207}.

\end{thebibliography}
